\documentclass[sigconf,dvipsnames,natbib=true,anonymous=false]{acmart}

\AtBeginDocument{%
  \providecommand\BibTeX{{%
    \normalfont B\kern-0.5em{\scshape i\kern-0.25em b}\kern-0.8em\TeX}}}

\copyrightyear{2023}
\acmYear{2023}
\setcopyright{acmlicensed}\acmConference[KDD '23]{Proceedings of the 29th
ACM SIGKDD Conference on Knowledge Discovery and Data Mining}{August
6--10, 2023}{Long Beach, CA, USA}
\acmBooktitle{Proceedings of the 29th ACM SIGKDD Conference on Knowledge
Discovery and Data Mining (KDD '23), August 6--10, 2023, Long Beach, CA,
USA}
\acmPrice{15.00}
\acmDOI{10.1145/3580305.3599452}
\acmISBN{979-8-4007-0103-0/23/08}

\usepackage{amsthm}
\usepackage{algorithm}
\usepackage{algpseudocode}
\usepackage{amsmath}
\usepackage{bm}
\usepackage{color}
\usepackage{qtree}
\usepackage{graphicx}
\usepackage{multirow}
\usepackage{multicol}
\usepackage{subfigure}
\usepackage{url}
\usepackage{thmtools}

\usepackage{tabularx}
\usepackage{balance}    % balance the two columns of the paper in the last page
\usepackage{booktabs}   % professional tables. See http://cs.brown.edu/about/system/managed/latex/doc/booktabs.pdf
\usepackage{graphics}   % image files in figure (e.g., pdf, eps files)
\usepackage{url}        % formating a web url using \url{...}
\usepackage{pifont}     % adding special characters using \ding{...}. See http://willbenton.com/wb-images/pifont.pdf

\usepackage{xcolor}      % color text
\usepackage{xspace}
\usepackage{balance}
\usepackage{enumitem}
\newcommand{\ie}[0]{\textit{i.e.},\ }   % i.e., meaning "which is/means "
\newcommand{\eg}[0]{\textit{e.g.},\ }   % e.g., meaning "for example, "
\usepackage[labelfont={bf,small},textfont={bf,small}]{caption}
\usepackage{pifont}% http://ctan.org/pkg/pifont
\newcommand{\cmark}{\ding{51}}%
\newcommand{\xmark}{\ding{55}}%

\usepackage{pgfplots}
\newcommand{\eat}[1]{}

\pgfplotsset{compat=1.18} 
\begin{document}

\title{One-shot Joint Extraction, Registration and Segmentation of Neuroimaging Data}

% \author{Yao Su}
% \affiliation{%
% \institution{Worcester Polytechnic Institute}
% \city{Worcester}
% \state{MA}
% \country{USA}
% \postcode{01609}}
% \email{ysu6@wpi.edu}

% \author{Zhentian Qian}
% \affiliation{%
% \institution{Worcester Polytechnic Institute}
% \city{Worcester}
% \state{MA}
% \country{USA}
% \postcode{01609}}
% \email{zqian@wpi.edu}

% \author{Lei Ma}
% \affiliation{%
% \institution{Worcester Polytechnic Institute}
% \city{Worcester}
% \state{MA}
% \country{USA}
% \postcode{01609}}
% \email{lma5@wpi.edu}

% \author{Lifang He}
% \affiliation{%
% \institution{Lehigh University}
% \city{Bethlehem}
% \state{PA}
% \country{USA}
% \postcode{18015}}
% \email{lih319@lehigh.edu}

% \author{Xiangnan Kong}
% \affiliation{%
% \institution{Worcester Polytechnic Institute}
% \city{Worcester}
% \state{MA}
% \country{USA}
% \postcode{01609}}
% \email{xkong@wpi.edu}

\author{Yao Su}
\affiliation{%
\institution{Worcester Polytechnic Institute}
\city{}
\state{}
\country{}
\postcode{}}
\email{ysu6@wpi.edu}

\author{Zhentian Qian}
\affiliation{%
\institution{Worcester Polytechnic Institute}
\city{}
\state{}
\country{}
\postcode{}}
\email{zqian@wpi.edu}

\author{Lei Ma}
\affiliation{%
\institution{Worcester Polytechnic Institute}
\city{}
\state{}
\country{}
\postcode{}}
\email{lma5@wpi.edu}

\author{Lifang He}
\affiliation{%
\institution{Lehigh University}
\city{}
\state{}
\country{}
\postcode{}}
\email{lih319@lehigh.edu}

\author{Xiangnan Kong}
\affiliation{%
\institution{Worcester Polytechnic Institute}
\city{}
\state{}
\country{}
\postcode{}}
\email{xkong@wpi.edu}

\renewcommand{\shortauthors}{Yao Su, Zhentian Qian, Lei Ma, Lifang He, \& Xiangnan Kong}
%% No italicsjer

\begin{abstract}
Brain extraction, registration and segmentation are indispensable preprocessing steps in neuroimaging studies. The aim is to extract the brain from raw imaging scans ({\ie} extraction step), align it with a target brain image ({\ie} registration step) and label the anatomical brain regions ({\ie} segmentation step).
Conventional studies typically focus on developing separate methods for the extraction, registration and segmentation tasks in a supervised setting.
The performance of these methods is largely contingent on the quantity of training samples and the extent of visual inspections carried out by experts for error correction.
Nevertheless, collecting voxel-level labels and performing manual quality control on high-dimensional neuroimages (\eg 3D MRI) are expensive and time-consuming in many medical studies.
In this paper, we study the problem of one-shot joint extraction, registration and segmentation in neuroimaging data, which exploits only one labeled template image (\emph{a.k.a.} atlas) and a few unlabeled raw images for training.
We propose a unified end-to-end framework, called JERS, to jointly optimize the extraction, registration and segmentation tasks, allowing feedback among them.
Specifically, we use a group of extraction, registration and segmentation modules to learn the extraction mask, transformation and segmentation mask, where modules are interconnected and mutually reinforced by self-supervision.
Empirical results on real-world datasets demonstrate that our proposed method performs exceptionally in the extraction, registration and segmentation tasks.

\begin{figure}[t]
  \centering
  \includegraphics[width=\linewidth]{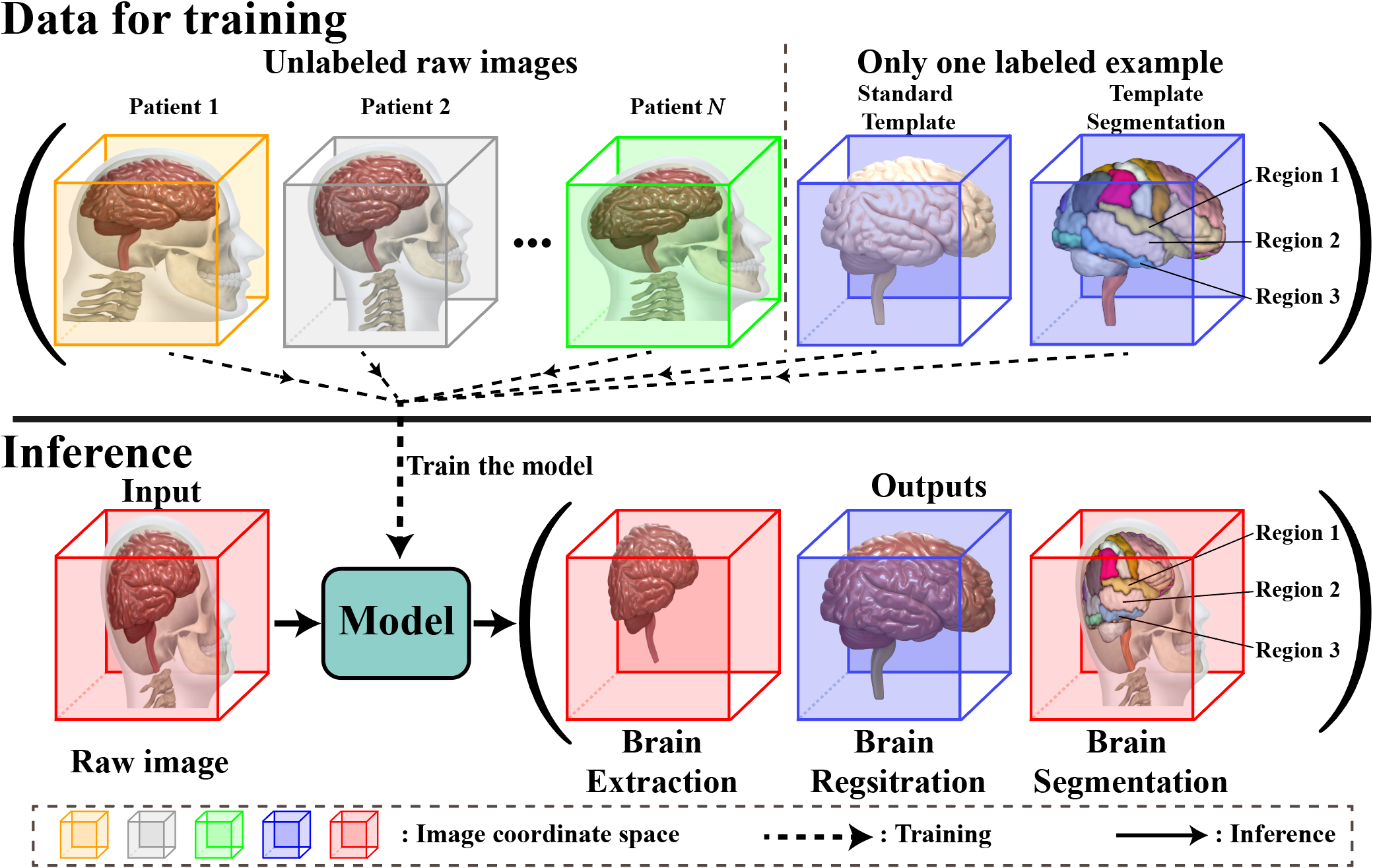}
  \vspace{-15pt}
  \caption{
The problem of one-shot joint extraction, registration and segmentation in neuroimaging data. Given a set of unlabeled raw images of the patients' heads, a standard template  image of the brain and the segmentation labels of the template image, the goal is to train a model to perform brain extraction, registration and segmentation tasks simultaneously on raw images of new patients' heads.
  }
  \label{fig:intro}
  \vspace{-15pt}
\end{figure}

\end{abstract}



% \ccsdesc[500]{Information systems~Data mining}
% \ccsdesc[500]{Computing methodologies~Neural networks}



\maketitle

% Introduction Section
\vspace{-6pt}
\section{Introduction}
\label{sec:intro}
\textbf{Background.} Brain extraction (\emph{a.k.a.} skull stripping), registration and segmentation serve as preliminary yet indispensable steps in many neuroimaging studies, such as anatomical and functional analysis~\cite{bai2017unsupervised,wang2017structural,yang2015structural,papalexakis2014good}, brain networks discovering~\cite{liu2017collective, liu2017unified, lee2017identifying, yin2018coherent, yin2020gaussian,dai2020recurrent}, multi-modality fusion~\cite{cai2018deep, 10.1145/3534678.3539301}, diagnostic assistance~\cite{sun2009mining,huang2011brain}, and brain region studies~\cite{chen2018voxel,lee2020deep}.
The brain extraction targets the removal of non-cerebral tissues (\eg skull, dura, and scalp) from a patient's head scan; the registration step aims to align the extracted brain with a standard brain template; the segmentation step intends to label the anatomical brain regions in the raw imaging scan. These three tasks serve as crucial preprocessing steps in many neuroimaging studies.
For example, in brain functional and anatomical analysis, upon extracting and aligning the brain, the interference of non-cerebral tissues, imaging modalities, and viewpoints can be eliminated, thereby enabling accurate quantification of shifts in the shape, size, and signal; and labeled anatomical brain regions (\eg frontal lobe, cerebellum, etc.) can be used to guide the structural diagnosis.
In Alzheimer's disease diagnosis, the brain across subjects needs to be first extracted from raw brain imaging scans and then aligned with a standard template to counteract inter-individual variations and perform brain function analysis (\eg discovering the brain network connectivity). Meanwhile, the intra-individual structural lesions (\eg brain atrophy) across different pathological stages need to be monitored in anatomical analysis (\eg identify the corresponding anatomical brain region and measure its alteration of brain volume). These essential processing steps help doctors make a comprehensive and accurate diagnosis.

\begin{figure}[t]
    \centering

    \subfigure[\textbf{
        Separate extraction \cite{smith2002fast,cox1996afni,shattuck2002brainsuite,segonne2004hybrid} + separate registration \cite{balakrishnan2018unsupervised,zhao2019recursive} + separate segmentation~\cite{jaderberg2015spatial}
        }]{
        \includegraphics[width=0.98\linewidth]{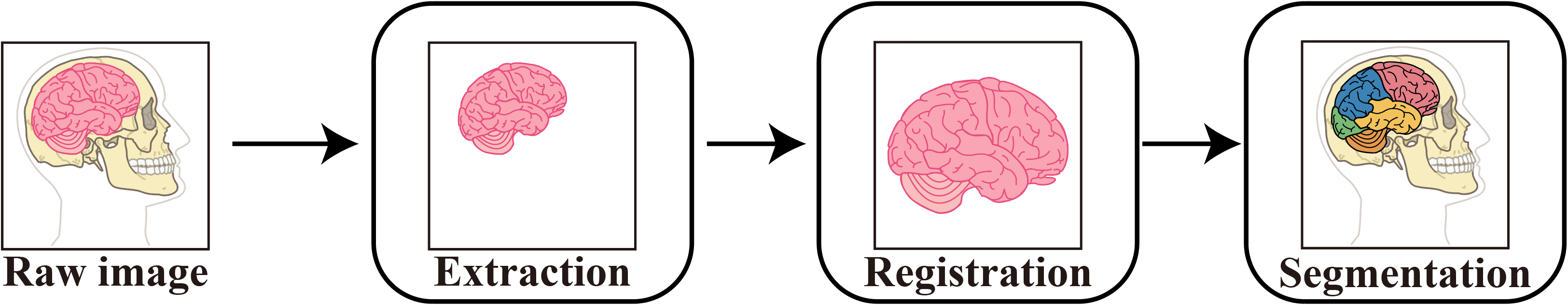}
        \label{fig:family 2}
    }
     \\[-0.8ex]
    \subfigure[\textbf{Joint extraction and registration~\cite{su2022ernet} + separate segmentation~\cite{jaderberg2015spatial}
    }]{
        \includegraphics[width=0.98\linewidth]{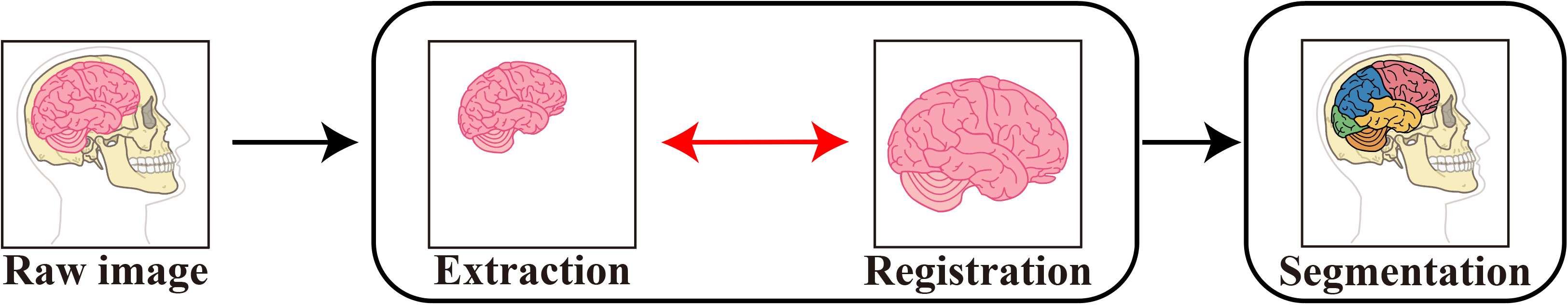}
        \label{fig:family 3}
            }
     \\[-0.8ex]
    \subfigure[\textbf{Separate extraction~\cite{smith2002fast,cox1996afni,shattuck2002brainsuite,segonne2004hybrid} + joint registration and segmentation~\cite{qiu2021u, xu2019deepatlas,he2020deep}}]{
        \includegraphics[width=0.98\linewidth]{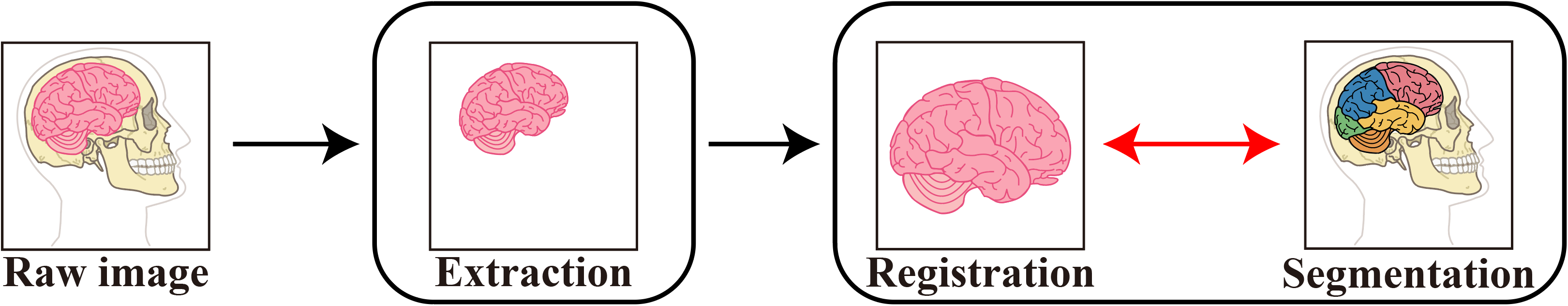}
        \label{fig:family 4}
            }
             \\[-1.2ex]
    \subfigure[\textbf{Joint extraction, registration and segmentation (ours)}]{
        \includegraphics[width=0.98\linewidth]{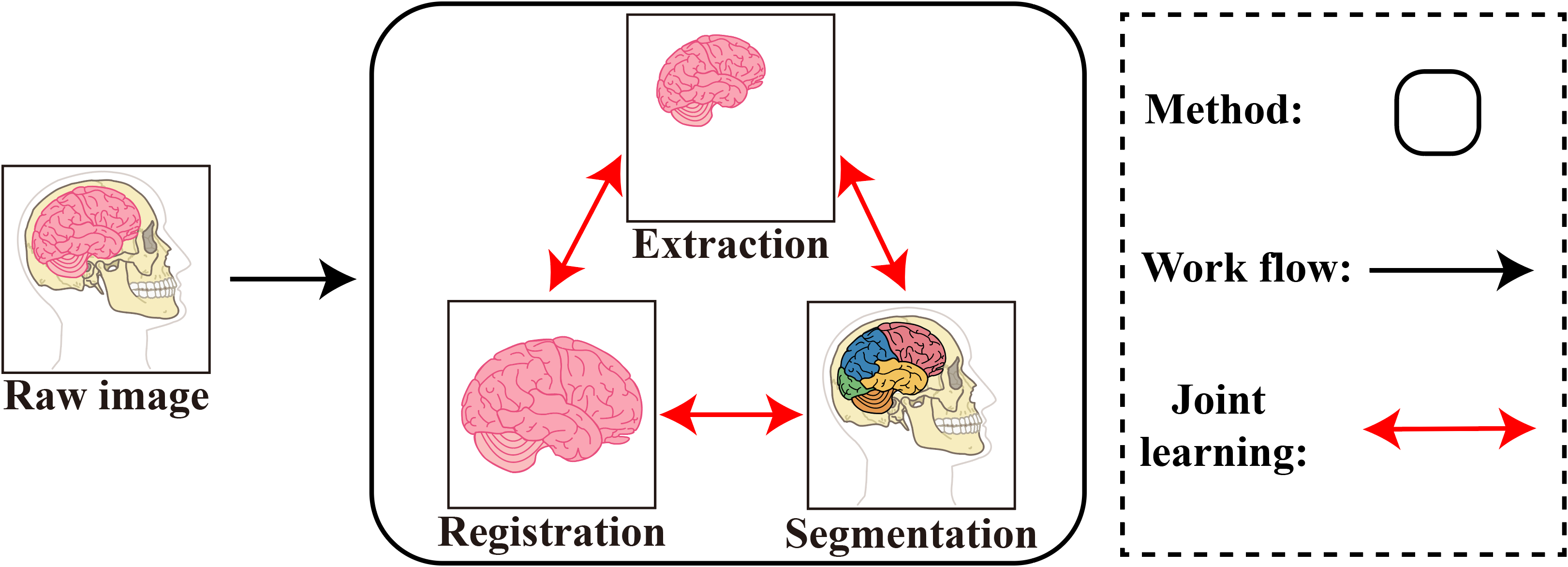}
        \label{fig:family 5}
        
    }
    \vspace{-12pt}
    \caption{
    Related works in one-shot brain extraction, registration and segmentation.
    }
    \vspace{-20pt}
    \label{fig: family}
\end{figure}

\textbf{State-of-the-Art.} 
The literature extensively explores brain extraction, registration, and segmentation problems~\cite{kleesiek2016deep,lucena2019convolutional,sokooti2017nonrigid, dai2020dual, akkus2017deep, chen2019learning, kamnitsas2017efficient}. Conventional approaches primarily emphasize the development of separate methods for extraction~\cite{kleesiek2016deep,lucena2019convolutional}, registration~\cite{sokooti2017nonrigid, dai2020dual}, and segmentation~\cite{akkus2017deep, chen2019learning, kamnitsas2017efficient} under supervised settings.
However, within the domain of medical studies,  the process of obtaining annotations for brain location, image transformations, and segmentation is often accompanied by significant expenses, necessitating expertise, and substantial time, especially when dealing with high-dimensional neuroimages (\eg 3D MRI).
To overcome this limitation, recent works~\cite{smith2002fast,cox1996afni,shattuck2002brainsuite,segonne2004hybrid,balakrishnan2018unsupervised,zhao2019recursive} introduce a three-step approach for one-shot extraction, registration and segmentation by using automated brain extraction tools~\cite{smith2002fast,cox1996afni,shattuck2002brainsuite,segonne2004hybrid}, unsupervised registration and segmentation models with direct warping~\cite{balakrishnan2018unsupervised,zhao2019recursive,jaderberg2015spatial}, as shown in Figure~\ref{fig:family 2}. However, these approaches often rely on manual quality control to correct intermediate results before performing subsequent tasks, which is time-consuming, labor-intensive, and subject to variability, thus hampering overall efficiency and performance.
More recently, joint extraction-registration method~\cite{su2022ernet} and joint registration-segmentation methods~\cite{qiu2021u, xu2019deepatlas,he2020deep} are introduced to solve the problem in a two-stage design, as shown in Figure~\ref{fig:family 3} and Figure~\ref{fig:family 4}. However, partial joint learning neglects the potential relationship among all tasks and negatively impacts overall performance.

\textbf{Problem Definition.} This paper investigates the problem of one-shot joint brain extraction, registration, and segmentation, as shown in Figure~\ref{fig:intro}. The goal is to capture the connections among three tasks to mutually boost their performance in a one-shot training scenario. Notably, the extraction, registration and segmentation labels of the raw image are not available. We expect to perform the three tasks simultaneously with only one labeled template. 

\textbf{Challenges.} Despite its value and significance, the problem of one-shot joint extraction, registration and segmentation has not been studied before and is very challenging due to its unique characteristics listed below:

\textbullet  \  
\textit{Lack of labels for extraction:} Traditional learning-based extraction methods rely on a substantial number of training samples with accurate ground truth labels. However, collecting voxel-level labels for high-dimensional neuroimaging data is a resource-intensive and time-consuming endeavor.

\textbullet  \  
\textit{Lack of labels for registration:} Obtaining the accurate ground truth transformation between raw and template images poses significant challenges. While unsupervised registration methods~\cite{balakrishnan2018unsupervised,zhao2019recursive} optimize transformation parameters by maximizing image similarity, their effectiveness is contingent upon the prior removal of non-brain tissue from the raw image. Failing to do this may lead to erroneous transformations, rendering the registration invalid.

\textbullet  \  
\textit{Lack of labels for segmentation:} Collecting the voxel-level segmentation labels is also difficult. Although we provide a template with its segmentation labeled (in template image space), the segmentation (in raw image space) of the raw image is not available.

\textbullet  \  
\textit{Dependencies among extraction, registration and segmentation:} Conventional research typically treats extraction, registration, and segmentation tasks separately. However, these tasks exhibit a high degree of interdependency. The accuracy of registration and segmentation tasks heavily relies on the extraction task. The registration process assists the extraction task in capturing cerebral/non-cerebral information from raw and template images, and providing segmentation labels for guiding the segmentation task. The segmentation task can inversely force the extraction and registration tasks to provide precise results. Thus, a holistic solution is required to effectively manage the interdependencies among these tasks.

%---------------------
% Figure 2

%---------------------

\textbf{Proposed Method.} To address the aforementioned challenges, we propose JERS, a unified end-to-end framework for joint brain extraction, registration, and segmentation. Figure~\ref{fig: family} showcases a comparison between our method and state-of-the-art approaches. Specifically, JERS contains a group of extraction, registration and segmentation modules, where the extraction module gradually eliminates the non-brain tissue from the raw image, producing an extracted brain image; the registration module incrementally aligns the extracted image with the template and warps the template's segmentation label in the raw image space to guide the segmentation module; the segmentation module generates a segmentation label for the raw image and provides feedback to extraction and registration modules. These three modules help each other to boost extraction, registration and segmentation performance simultaneously. By bridging these three modules end-to-end, we achieve a joint optimization with only one labeled template assistance.

%Extensive experiments on public brain MRI datasets demonstrate that our proposed method significantly outperforms state-of-the-art approaches in extraction, registration, and segmentation accuracy.

Upon evaluation on public brain MRI datasets, our proposed method significantly outperforms state-of-the-art techniques in extraction, registration, and segmentation accuracy.

% %-----------------------------------------------
% % Problem Definition Section
\section{PRELIMINARIES} 

In this section, we begin by introducing the relevant concepts and notations. We then proceed to formally define the problem of one-shot joint brain extraction, registration, and segmentation.

\subsection{Notations and Definitions}

\noindent \textbf{Definition 1 (Source, target and target segmentation mask).}
Suppose we are given a training dataset $\mathcal{D} = \left\{\{\mathbf{S}_i\}_{i=1}^{Z}, (\mathbf{T}, \mathbf{B})\right\}$ that consists of $Z$ source images $\mathbf{S}_{i} \in \mathbb{R}^{W \times H \times D}$, and a pair of target image $\mathbf{T} \in \mathbb{R}^{W \times H \times D}$ and its corresponding segmentation mask $\mathbf{B} \in \{0, 1\}^{C \times W \times H \times D}$. Here, the source image $\mathbf{S}_i$ is the raw MRI scan of a patient's head, the target $\mathbf{T}$ is a standard template of the brain, and the segmentation mask $\mathbf{B}$ is the one-hot encoding of the target $\mathbf{T}$ segmentation. $W$, $H$, and $D$ denote the width, height and depth dimensions of the 3D images, $C$ denotes the number of anatomical labels (\eg the number of labelled brain regions). 
To ensure simplicity, we make the assumption that the source and target images are resized to the same dimension, denoted as $W \times H \times D$. Next, we omit the subscript $i$ of $\mathbf{S}_{i}$ for ease of notation.

%For simplicity, we assume that the source and target images are resized to the same dimension, \ie $W \times H \times D$. 
%In the following discussion, we omit the subscript $i$ of $\mathbf{S}_{i}$ for ease of notation. 

\noindent \textbf{Definition 2 (Brain extraction mask).}
Brain extraction mask $\mathbf{M} \in \{0,1\}^{W \times H \times D}$ is a binary tensor of identical dimensions to the source image $\mathbf{S}$.
It represents cerebral tissues in $\mathbf{S}$ with a value of 1 and non-cerebral tissues with 0.
The extracted image $\mathbf{E} = \mathbf{S} \circ \mathbf{M}$ is obtained by applying the $\mathbf{M}$ on $\mathbf{S}$ via a element-wise product $\circ$.

\noindent \textbf{Definition 3 (Affine transformation and warped image).}
In order to maintain generality, we consider the transformation in the registration task to be affine-based. 
%Without loss of generality, we consider that the transformation in the registration task is affine-based. 
Extending this work to encompass other types of registration, such as nonlinear/deformable registration, is straightforward.
%Extending this work to other types of registration, such as nonlinear/deformable ones is straightforward.
The affine transformation parameters $\mathbf{a} \in \mathbb{R}^{12} $ is a vector used to parameterized an 3D affine transformation matrix $\mathbf{A} \in \mathbb{R}^{4 \times 4} $. The warped image $\mathbf{W} = \mathcal{T}\left(\mathbf{E},\mathbf{a}\right)$ results from applying the affine transformation on the extracted image $\mathbf{E}$, where $\mathcal{T}(\cdot, \cdot)$ denotes the affine transformation operator. 
At the voxel level, the relationship between $\mathbf{W}$ and $\mathbf{E}$ can be expressed as:
%The following relationship holds for $\mathbf{W}$ and $\mathbf{E}$ on the voxel level: 
\begin{equation}
\label{equ:voxel_value}
\mathbf{W}_{xyz} = \mathbf{E}_{x'y'z'},
\end{equation}
where the correspondences between coordinates $x,y,z$ and $x',y',z'$ are determined on the affine transformation matrix $\mathbf{A}$:
\begin{equation}
\begin{bmatrix}
x'\\
y'\\
z' \\
1
\end{bmatrix} = \mathbf{A}\begin{bmatrix}
x\\
y\\
z \\
1
\end{bmatrix} = \begin{bmatrix}
a_{1} & a_{2} & a_{3} & a_{4}\\
a_{5} & a_{6} & a_{7} & a_{8}\\
a_{9} & a_{10} & a_{11} & a_{12}\\
0 & 0 & 0 & 1
\end{bmatrix} \begin{bmatrix}
x\\
y\\
z \\
1
\end{bmatrix}.
\end{equation}

\noindent \textbf{Definition 4 (Source segmentation mask).}
Brain segmentation mask $\mathbf{R} \in \{0,1\}^{C \times W \times H \times D}$ is a binary tensor with the first dimension being the number of anatomical labels and the rest dimensions identical to the source image $\mathbf{S}$. Every point in the source image $\mathbf{S}$ is densely labeled according to its anatomical structure, encoded in a one-hot vector at the corresponding coordinate of $\mathbf{R}$.

\begin{figure}[t]
  \centering
  \includegraphics[width=\linewidth]{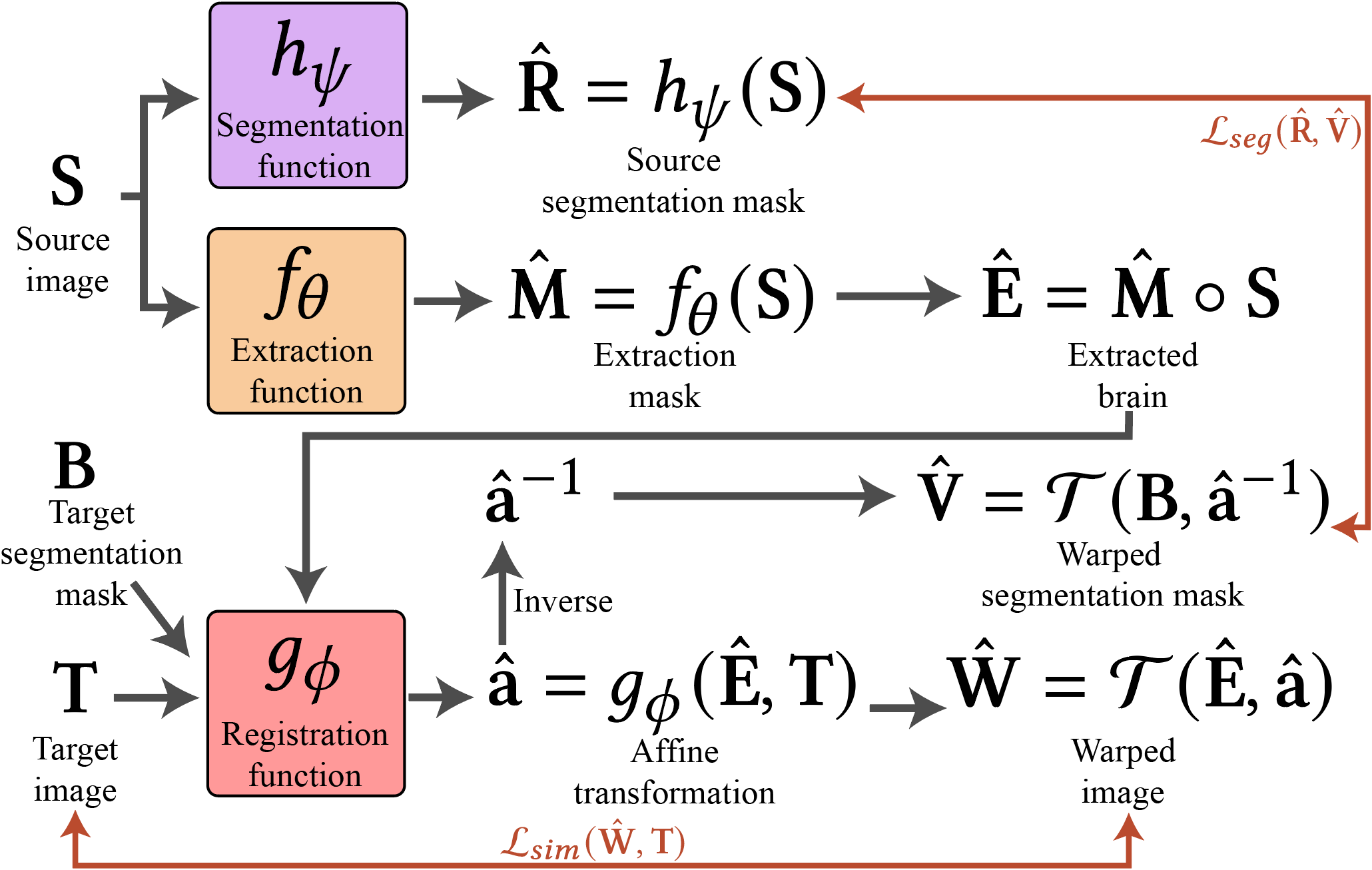}
  \vspace{-12pt}
  \caption{A demonstration of extraction, registration and segmentation functions.}
  \label{fig:formulation}
  \vspace{-15pt}
\end{figure}
\subsection{Problem Formulation}

The goal of joint brain extraction, registration, and segmentation is to collectively learn the extraction function $f_{\theta}: \mathbb{R}^{W \times H \times D} \rightarrow \mathbb{R}^{W \times H \times D}$, the registration function $g_{\phi}: \mathbb{R}^{W \times H \times D}\times \mathbb{R}^{W \times H \times D} \rightarrow \mathbb{R}^{12}$, and the segmentation function $h_{\psi}: \mathbb{R}^{W \times H \times D} \rightarrow \mathbb{R}^{C \times W \times H \times D}$, as shown in Figure~\ref{fig:formulation}. Specifically, the extraction function $f_{\theta}(\cdot)$ utilizes the source image $\mathbf{S}$ as input to predicts a brain extraction mask $\hat{\mathbf{M}} = f_{\theta}(\mathbf{S})$. The registration function $g_{\phi}(\cdot, \cdot)$ takes the extracted brain image $\hat{\mathbf{E}} = \hat{\mathbf{M}} \circ \mathbf{S}$ and the target image $\mathbf{T}$ to predict the affine transformation parameter $\hat{\mathbf{a}} = g_{\phi}(\hat{\mathbf{E}},\mathbf{T})$, then obtaining the warped image $\hat{\mathbf{W}} = \mathcal{T}(\hat{\mathbf{E}},\hat{\mathbf{a}})$. 
The warped segmentation mask $\hat{\mathbf{V}} = \mathcal{T}(\mathbf{B},\hat{\mathbf{a}}^{-1})$ is generated by warping the target segmentation mask $\mathbf{B}$ using inversed affine transformation $\hat{\mathbf{a}}^{-1}$. Finally, the segmentation function  $h_{\psi}(\cdot)$ takes the source image $\mathbf{S}$ as input to predict a source brain segmentation mask $\hat{\mathbf{R}} = h_{\psi}(\mathbf{S})$.
The optimal parameter $\theta^*$, $\phi^*$ and $\psi^*$ can be found by solving the following optimization problem:
\begin{equation}
\begin{split}
\label{equ:goal_training}
\theta^{*},\phi^{*},\psi^{*} &=\underset{\theta,  \phi,\psi}{\arg \min } \hspace{-3pt} \sum_{\left(\mathbf{S},  \mathbf{T}, \mathbf{B}\right)\in \mathcal{D}}\left[ \mathcal{L}_{sim} \left( \hat{\mathbf{W}}, \mathbf{T} \right) + \lambda \mathcal{L}_{seg} \left( \hat{\mathbf{R}}, \hat{\mathbf{V}}  \right)\right],
\end{split}
\end{equation}
where the image pair $(\mathbf{S},\mathbf{T},\mathbf{B})$ is sampled from the training dataset $\mathcal{D}$. $\mathcal{L}_{sim}(\cdot, \cdot)$ is image dissimilarity criteria, \eg mean square error and $\mathcal{L}_{seg}(\cdot, \cdot)$ is segmentation dissimilarity criteria, \eg cross entropy error. These two criteria guide a joint optimization of extraction, registration and segmentation functions, allowing feedback among them.

To the best of our knowledge, this work is the first endeavor in finding an optimal solution for the one-shot joint brain image extraction, registration, and segmentation problem. Our approach eliminates the necessity of labeling the brain extraction masks, transformation, and segmentation masks of the source image. We only require one pair of a target image and its corresponding segmentation mask to guide the training, as opposed to other fully supervised methods~\cite{kleesiek2016deep,lucena2019convolutional,sokooti2017nonrigid, dai2020dual, akkus2017deep, chen2019learning, kamnitsas2017efficient}.

% %-----------------------------------------------
% % Method Section

\definecolor{myorange}{RGB}{249, 203, 156}
\definecolor{myviolet}{RGB}{218, 175, 244}
\definecolor{mypink}{RGB}{250, 127, 111}

\begin{figure*}[t]
  \centering
  \includegraphics[width=1.0\linewidth]{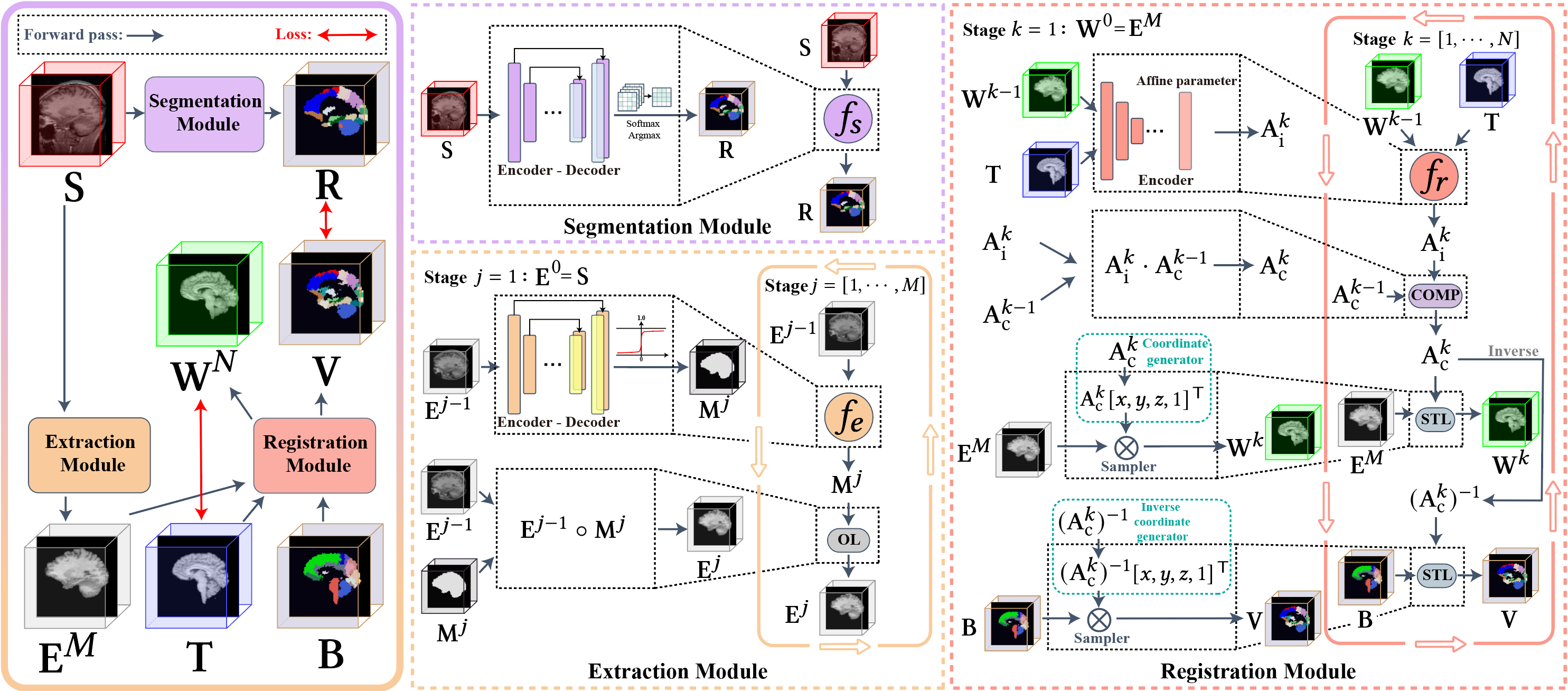}
    \vspace{-12pt}
  \caption{An overview of proposed JERS. \textcolor{myorange}{Extraction module} takes the raw source image $\mathbf{S}$ as input, and gradually produces the extracted brain image $\mathbf{E}^{M}$ after $M$ stages of extraction.
  The final extracted image is $\mathbf{E}^{M}$.
  \textcolor{mypink}{Registration module} takes the extracted brain image $\mathbf{E}^{M}$, target image $\mathbf{T}$ and target segmentation mask $\mathbf{B}$ as inputs, and incrementally aligns $\mathbf{E}^{M}$ with $\mathbf{T}$ through $N$ stages of registration. Then, it generates the warped segmentation mask $\mathbf{V}$ by inversely transforming the target segmentation mask $\mathbf{B}$.
  \textcolor{myviolet}{Segmentation module} takes the raw source image $\mathbf{S}$ as input, and output the brain segmentation mask $\mathbf{R}$.
  Two loss terms couple all modules together, allowing them to perform joint learning.
  The final output of JERS is the extracted brain image $\mathbf{E}^{M}$ constituting only cerebral tissues, the warped image $\mathbf{W}^{N}$ aligning with the target image $\mathbf{T}$, and the brain segmentation mask $\mathbf{R}$ indicating anatomical regions of the source image $\mathbf{S}$.
    }
  \label{fig:network}
  \vspace{-12pt}
\end{figure*}

\section{Our Approach}
\label{sec:method}
\noindent\textbf{Overview.} Figure~\ref{fig:network} presents an overview of the proposed JERS framework for the one-shot joint brain extraction, registration, and segmentation problem. Our method is an end-to-end deep neural network consisting of three main branches: 1) \emph{Extraction Module} takes the raw source image $\mathbf{S}$ as input, and progressively produces the extracted brain image $\mathbf{E}^{M}$ after $M$ stages of extraction; 2) \emph{Registration Module} takes the extracted brain image $\mathbf{E}^{M}$, target image $\mathbf{T}$ and target segmentation mask $\mathbf{B}$ as inputs, and incrementally aligns $\mathbf{E}^{M}$ with $\mathbf{T}$ through $N$ stages of registration. Then, it generates the warped segmentation mask $\mathbf{V}$ (\ie segmentation mask in the source image space) by conducting inverse transformation on the target segmentation mask $\mathbf{B}$; 3) \emph{Segmentation Module} takes the raw source image $\mathbf{S}$ as inputs, and output the brain segmentation mask $\mathbf{R}$. The final output of our network is the extracted brain image $\mathbf{E}^{M}$ constituting only cerebral tissues, the warped image $\mathbf{W}^{N}$ aligning with the target image, and the brain segmentation mask $\mathbf{R}$ indicating anatomical regions of the source image.

\noindent \textbf{Key Insight.}  Considering the highly correlated characteristics of the extraction, registration, and segmentation tasks, the design of our method revolves around the following four positive mechanisms among the three modules:

\textbullet  \  The unalignable non-brain tissue revealed in the registration module guides the refinement process in the extraction module.

\textbullet  \  The registration module benefits from accurate brain extraction generated in the extraction module.

\textbullet  \  The anatomical structure estimated by the segmentation module can provide auxiliary information for the extraction module to remove non-cerebral tissues and for the registration module to find regional correspondences. 

\textbullet  \  The segmentation module relies on the extraction and registration modules to generate the corresponding segmentation mask in the source image space.

Hence, we encode the above four positive mechanisms in our loss function design and train our network in an end-to-end fashion. Specifically, our loss function includes two perceptual losses: the image similarity term $\mathcal{L}_{sim}( \mathbf{W}^{N}, \mathbf{T})$ and the segmentation loss term $\mathcal{L}_{seg}( \mathbf{R}, \mathbf{V})$.
The image similarity loss $\mathcal{L}_{sim}(\cdot, \cdot)$ captures the first and second positive mechanisms, which is affected by extraction and registration modules. 
The segmentation loss $\mathcal{L}_{seg}(\cdot, \cdot)$ embodies the third and fourth positive mechanisms, which is related to all modules.
We jointly optimize the two loss terms to guide the learning of extraction, registration, and segmentation modules.
Next, we introduce the details of each module and the training process.

\subsection{Extraction Module}
\label{sec: Extration Net}
For the extraction module, we have adopted the multi-stage design paradigm, a strategy that has proven effective in previous works~\cite{su2022ernet, zhao2019recursive, feng2021recurrent,tang2019multi,su2022abn}. This design allows for a progressive refinement in the removal of non-cerebral tissues, culminating in an image with only cerebral tissues at the final stage.
Following the multi-stage design of~\cite{su2022ernet}, the extraction module incorporates $M$ extraction stages, and each stage $j$ contains two key components as follows.

\subsubsection{Extraction Network: $f_{e}$}
\label{sec: fe}
The extraction network $f_{e}(\cdot)$ acts as an eliminator, intended to sequentially eradicate non-cerebral elements from source image $\mathbf{S}$, aiming to retain only brain tissue in the image obtained at the final stage. With each stage $j$, it forms an extraction mask $\mathbf{M}^{j}$ , which is utilized to eliminate non-cerebral tissues in the preceding extracted brain image $\mathbf{E}^{j-1}$. 
Specifically, we employ the 3D U-Net \cite{ronneberger2015u} as the base network to learn $f_{e}(\cdot)$.
When conducting inference, the output $\mathbf{M}^{j}$ is binarized by a Heaviside step function.
To facilitate effective gradient backpropagation during the training phase, we employ a Sigmoid function with a large slope parameter to approximate the Heaviside step function. 
$f_{e}(\cdot)$ employs a shared-weight scheme, which means $f_{e}(\cdot)$ is repeatedly applied across all stages with the same set of parameters.
The process can be formally expressed as:

\begin{equation}
\mathbf{M}^{j}=f_{e}\left(\mathbf{E}^{j-1}\right),
\end{equation}
where $\mathbf{M}^{j}$ is the outputted brain mask of the $j$-th stage for $j = [1, \cdots, M]$ and $\mathbf{E}^{0} = \mathbf{S}$.

\subsubsection{Overlay Layer: $OL$}
The overlay layer serves to eradicate any residual non-cerebral tissues by applying the current brain mask $\mathbf{M}^{j}$ to the preceding extracted image $\mathbf{E}^{j-1}$. The updated extraction is $ \mathbf{E}^{j} = \mathbf{E}^{j-1}\circ \mathbf{M}^{j}$, where $\circ$ denotes the operation of element-wise multiplication.

\subsection{Registration Module}
Similar to the extraction module introduced in Section \ref{sec: Extration Net}, we execute a multi-stage paradigm~\cite{su2022ernet, zhao2019recursive, feng2021recurrent,tang2019multi,su2022abn} to tackle the registration task. 
Following the multi-stage
settings of~\cite{su2022ernet}, the module is composed of $N$ cascading stages, with each stage $k$ holding the key components as follows.

\subsubsection{Registration Network: $f_{r}$}
The purpose of the registration network $f_{r}(\cdot, \cdot)$ is to slowly adapt the extracted brain image to better match the target image. Each stage $k$ generates a current affine transformation $\mathbf{A}_\text{i}^{k}$ based solely on the prior warped image $\mathbf{W}^{k-1}$ and the target image $\mathbf{T}$.
Similar to the extraction network $f_e$ in Section~\ref{sec: fe}, a 3D CNN-based encoder is used to learn $f_{r}(\cdot, \cdot)$ and a shared weight design is employed across stages using identical parameters. As a formal expression,
\begin{equation}
\mathbf{A}_\text{i}^{k}=f_{r}\left(\mathbf{W}^{k-1}, \mathbf{T}\right),
\end{equation}
where $\mathbf{A}_\text{i}^{k}$ represents the output affine transformation of the $k$-th stage for $k = [1, \cdots, N]$ and $\mathbf{W}^{0} = \mathbf{E}^{M}$.

\subsubsection{Composition Layer: $\text{COMP}$}
For each stage $k$, upon determining the current affine transformation $\mathbf{A}_\text{i}^{k}$ through $f_{r}(\cdot, \cdot)$, we combine all preceding transformations: $\mathbf{A}_\text{c}^{k}=\mathbf{A}_\text{i}^{k} \cdot \mathbf{A}_\text{c}^{k-1}$, where $\cdot$ denoting the matrix multiplication operation. 
Thus, the combined transformation $\mathbf{A}_\text{c}^{k}$ can always be performed on the extracted image $\mathbf{E}^{M}$ to preserve the sharpness of the warped images~\cite{su2022abn}.
For $k=1$, the initial affine transformation $\mathbf{A}_\text{c}^{0}$ is designated as an identity matrix, indicating no displacement.

\subsubsection{Spatial Transformation Layer: $\text{STL}$}
\label{sec: STL}
A crucial process in image registration involves reconstructing the warped image $\mathbf{W}^{k}$ from the extracted brain image $\mathbf{E}^{M}$ using the affine transformation operator. Utilizing the combined transformation $\mathbf{A}_\text{c}^{k}$, we introduce a spatial transformation layer (STL) that resamples voxels from the extracted image to produce the warped image through $\mathbf{W}^{k} = \mathcal{T}(\mathbf{E}^{M}, \mathbf{A}_\text{c}^{k})$.
Given the affine transformation operator depicted in Eq.~(\ref{equ:voxel_value}), we hold
\begin{equation}
     \mathbf{W}_{xyz}^{k} = \mathbf{E}_{x'y'z'}^{M} \hspace{1pt},
     \label{equ:voxel_value_k}
\end{equation}
where $[x', y', z', 1]^\top = \mathbf{A}_\text{c}^{k}[x, y, z, 1]^\top $. To assure successful gradient propagation during this procedure, we adopt a differentiable transformation founded on trilinear interpolation proposed by~\cite{jaderberg2015spatial}.

\subsection{Inverse Warping and Segmentation Module}
This section introduces the two key designs to solve the segmentation task for the input source image $\mathbf{S}$: 1) \emph{Inverse Warping} takes the target segmentation mask $\mathbf{B}$ and the composed affine transformation $\mathbf{A}_\text{c}^{N}$ as inputs, and generates a segmentation mask $\mathbf{V}$ in the source image space; 2) \emph{Segmentation Network} takes the source image $\mathbf{S}$ as input, and predicts the brain segmentation mask $\mathbf{R}$ for $\mathbf{S}$.  

\subsubsection{Inverse Warping}
In the training dataset, only the target image $\mathbf{T}$ is labeled with its segmentation mask $\mathbf{B}$. In order to generate the segmentation label for guiding the segmentation network, we warp the target image segmentation make $\mathbf{B}$ into the source image space by applying the inverse of $\mathbf{A}_\text{c}^{N}$:
\begin{equation}
    \mathbf{V}_{cxyz} = \mathbf{B}_{cx'y'z'}, \forall  c \in \{0, 1 ,\ldots, C-1\}
\end{equation}
where $[x', y', z', 1]^\top = (\mathbf{A}_\text{c}^{N})^{-1}[x, y, z, 1]^\top $, $c$ is the index for anatomical class. 
Same as the $\text{STL}$ in Section~\ref{sec: STL}, we apply a differentiable transformation based on trilinear interpolation. 

\subsubsection{Segmentation Network: $f_s$}
The segmentation network is tasked to predict a segmentation mask for the source image $\mathbf{S}$ that matches the synthesized segmentation mask $\mathbf{V}$. Similar to the extraction network discussed in Section \ref{sec: fe}, we employ the popular 3D U-Net as the base network to learn $f_s(\cdot)$. Formally, we have:
\begin{equation}
    \mathbf{R} = f_s(\mathbf{S}).
\end{equation}

\subsection{End-to-End Training}
\label{section:end-to-end training}

We train our JERS model by minimizing the following objective function
\begin{equation}
\label{eq:loss}
\theta^*, \phi^*, \psi^* = \underset{\theta, \phi, \psi}{\min} \mathcal{L}_{sim}(\mathbf{W}^N, \mathbf{T}) + \lambda\mathcal{L}_{seg}(\mathbf{R}, \mathbf{V})  + \eta \sum_{j=1}^{M} \mathcal{R}(\mathbf{M}^{j}),
\end{equation}
where $\theta, \phi, \psi$ are the parameters for the extraction, registration, and segmentation networks, respectively. $\mathcal{L}_{sim}(\cdot, \cdot)$ is the image  similarity loss, $\mathcal{L}_{seg}(\cdot, \cdot)$ is the segmentation loss, $\mathcal{R}$ is the extraction regularization term, and $\lambda, \eta$ are positive trade-off weights. 

The image similarity loss $ \mathcal{L}_{sim}(\cdot, \cdot)$ is a loss function measuring the similarity between the final warped image $\mathbf{W}^{N}$ and the target image $\mathbf{T}$:
\begin{equation}
    \mathcal{L}_{sim}\big(\mathbf{W}^{N}, \mathbf{T}\big) =  \mathcal{L}_{sim}\Big(\mathcal{T}\big(\mathbf{E}^{M}, \mathbf{A}_\text{c}^{N}\big), \mathbf{T}\Big). 
\end{equation}
We employ  the prevailing negative local cross-correlation loss for $\mathcal{L}_{sim}(\cdot, \cdot)$, which is resistant to variations in voxel intensity typically encountered across different scans and datasets~\cite{balakrishnan2018unsupervised, xu2019deepatlas, zhao2019recursive}.
The image similarity loss can guide the learning of both the extraction and registration modules.

The segmentation loss $\mathcal{L}_{seg}(\cdot, \cdot)$ is a loss function measuring the similarity between the predicted segmentation mask $\mathbf{R}$ and the warped segmentation label $\mathbf{V}$:
\begin{equation}
\mathcal{L}_{seg}(\mathbf{R}, \mathbf{V}) = \mathcal{L}_{seg}\left(f_s\big(\mathbf{S}\big), \mathcal{T}\Big(\mathbf{B}, \big(\mathbf{A}_\text{c}^{N}\big)^{-1}\Big)\right)
\end{equation}
Here we use the cross-entropy loss function. The segmentation loss serves multiple purposes and can guide the learning of the extraction, registration, and segmentation modules. 

Finally, to suppress the occurrence of other connected regions apart from the brain in the extraction masks, we define the extraction regularization loss:
\begin{equation}
    \mathcal{R}(\mathbf{M}^{j}) = \textstyle\sum_{x=1}^{W} \sum_{y=1}^{H}  \sum_{z=1}^{D}\|\nabla \mathbf{M}_{xyz}^{j}\|^{2}.
\end{equation}
This regularization term quantifies the edge strength of the predicted extraction mask $\mathbf{M}^{j}$, \ie the likelihood of a voxel being an edge voxel. Through the minimization of this term, we can suppress the occurrence of edges, subsequently producing a smooth extraction mask. In this scenario, we apply the $\ell_2$-norm of the first-order derivative of $\mathbf{M}^{j}$ as the regularization term.

By leveraging the differentiability in each component of this design, our model can be jointly and progressively optimized in an end-to-end fashion. This training approach empowers us to unearth a joint optimal solution for the aggregate tasks of brain extraction, registration, and segmentation. Thus our work stood in marked contrast to other works \cite{su2022ernet, xu2019deepatlas} that resort to alternative training of individual modules and can only reach a sub-optimal solution.

% %-----------------------------------------------
% % Experiment Section
%\section{Experiment and Analysis}
\section{Experiments}

\subsection{Datasets} \label{section:Dataset}

The effectiveness of our proposed method is evaluated across three public real-world 3D brain MRI datasets: 1) \emph{LPBA40}~\cite{shattuck2008construction} includes 40 raw T1-weighted 3D MRI scans, coupled with brain masks and 56 anatomical structures as segmentation ground truths; 2) \emph{CC359}~\cite{souza2018open} contains 359 raw T1-weighted 3D brain MRI scans and the brain masks. It also includes labeled white matter as the segmentation ground truth; 3) \emph{IBSR}~\cite{rohlfing2011image} consists of 18 scans with manual segmentation labels. But due to its small size, it serves only for the evaluation of the model trained on CC359. Brain masks and anatomical labels facilitate the accuracy evaluation for extraction and segmentation, respectively. Additional details can be found in Appendix~\ref{section: appendix Data Preprocessing}.

\subsection{Compared Methods}
\label{section:Compared Methods}

We outline the comparison of our JERS with several representative methodologies in brain extraction, registration and segmentation, as illustrated in Table~\ref{tab:methods}. Notably, there are no existing solutions that can seamlessly learn brain extraction, registration, and segmentation in an end-to-end framework. Hence, we designed three-stage pipeline for comparison, combining various extraction, registration, and segmentation methods. Baselines detailed in Appendix~\ref{section: appendix baselines settings}.

\noindent\textbullet\ \textit{Brain Extraction Tool (BET)} \cite{smith2002fast}: 
This technique within the FSL package uses deformable models for skull stripping.

\noindent\textbullet\ \textit{3dSkullStrip (3dSS)} \cite{cox1996afni}: This is a BET variant method within the AFNI package. It uses a spherical surface to perform skull stripping.

\noindent\textbullet\ \textit{Brain Surface Extractor (BSE)} \cite{shattuck2002brainsuite}: This method uses morphological operations and edge detection for brain extraction, which leverages anisotropic diffusion filtering and a Marr Hildreth edge detector to identify the brain boundary.

\noindent\textbullet\ \textit{FMRIB's Linear Image Registration Tool (FLIRT)} \cite{jenkinson2001global}: This is an automatic affine brain image registration tool in the FSL package.

\noindent\textbullet\ \textit{Advanced Normalization Tools (ANTs)} \cite{avants2009advanced}: This is a state-of-the-art medical image registration toolbox. We set the registration type and optimization as affine and cross-correlation metrics.

\noindent\textbullet\ \textit{VoxelMorph (VM)} \cite{balakrishnan2018unsupervised}: This is an unsupervised image registration method that uses a neural network to predict the transformation.

\noindent\textbullet\ \textit{Cascaded Registration Networks (CRN)} \cite{zhao2019recursive}:
This is an unsupervised multi-stage image registration method, which incrementally aligns the source image to the target image.  

\noindent\textbullet\ \textit{Directly Warping (DW)~\cite{jaderberg2015spatial}}:
It is an operation used to generate the segmentation mask by registration. The segmentation mask of the target image can be directly warped to the source image space by the completed registration.

\noindent\textbullet\ \textit{DeepAtlas \cite{xu2019deepatlas}}: This is a joint learning network for image registration and segmentation tasks.

\noindent\textbullet\ \textit{ERNet \cite{su2022ernet}}: This is an unsupervised learning method for joint extraction and registration.

\begin{table}[t]
    \centering
    \caption{Summary of compared methods.}
    \label{tab:methods}
    \vspace{-10pt}
    \resizebox{\linewidth}{!}{
    \begin{tabular}{lcccc}
    \toprule
    \textbf{Methods}& \textbf{Extraction}& \textbf{Registration}& \textbf{Segmentation}& \textbf{Deep learning}\\
    \midrule
    BET { \cite{smith2002fast}} & \cmark  & \xmark & \xmark & \xmark \\
    % BET$^*$ { \cite{smith2002fast}} & \cmark  & \xmark & \cmark & \xmark\\
    3dSkullStrip \cite{cox1996afni}& \cmark  & \xmark & \xmark & \xmark\\
    BSE \cite{shattuck2002brainsuite}& \cmark  & \xmark & \xmark & \xmark\\
    \midrule
    FLIRT \cite{jenkinson2001global} & \xmark  & \cmark & \xmark & \xmark\\
    ANTs \cite{avants2009advanced}& \xmark  & \cmark & \xmark & \xmark\\
    VM \cite{balakrishnan2018unsupervised}& \xmark  & \cmark & \xmark & \cmark\\
    CRN \cite{zhao2019recursive}& \xmark  & \cmark & \xmark & \cmark \\
    \midrule
    DW \cite{jaderberg2015spatial} & \xmark  & \xmark & \cmark & \xmark \\
    \midrule
    DeepAtlas \cite{xu2019deepatlas}& \xmark  & \cmark &
    \cmark & \cmark\\
    ERNet \cite{su2022ernet}& \cmark  & \cmark & \xmark & \cmark\\
    \midrule
    JERS (ours)& \cmark  & \cmark & \cmark & \cmark\\
    
    \bottomrule
    \end{tabular}
    }
    \vspace{-15pt}
\end{table}

\begin{table*}[t]
    \centering
    \caption{Results for brain extraction (Ext), registration (Reg) and segmentation (Seg) in different datasets. The results are reported as performance(mean $\pm$ std ) of extraction, registration and segmentation of each compared method.
    “$\uparrow$” point out “the larger the better”. 
    The best results are highlighted in bold.}
    \label{tab:main res}
    \vspace{-10pt}
    \resizebox{1.0\linewidth}{!}{
    \begin{tabular}{lccccccccccc}
    \toprule
    \multicolumn{3}{c}{\multirow{2}{*}{Methods}}                & \multicolumn{9}{c}{Datasets}   \\                                                    
    \cmidrule(lr){4-12}
    
    \multicolumn{3}{c}{}                                        & \multicolumn{3}{c}{LPBA40} & \multicolumn{3}{c}{CC359} & \multicolumn{3}{c}{IBSR}\\
    
    \cmidrule(lr){1-3}\cmidrule(lr){4-6}\cmidrule(lr){7-9}\cmidrule(lr){10-12}
    
    \multirow{2}{*}{Ext} & \multirow{2}{*}{Reg} & \multirow{2}{*}{Seg} & Ext  & Reg & Seg & Ext & Reg  & Seg & Ext & Reg & Seg \\   
                                    &     &    &
 Dice$_{\mathrm{ext}}$  $\uparrow$        & MI  $\uparrow$   &  Dice$_{\mathrm{seg}}$ $\uparrow$ 
 & Dice$_{\mathrm{ext}}$ $\uparrow$       & MI  $\uparrow$  & Dice$_{\mathrm{seg}}$ $\uparrow$         & Dice$_{\mathrm{ext}}$ $\uparrow$       & MI  $\uparrow$    & Dice$_{\mathrm{seg}}$ $\uparrow$  \\

     \midrule

    BET \cite{smith2002fast} & FLIRT \cite{jenkinson2001global} & DW~\cite{jaderberg2015spatial} & 0.935  $\pm$   0.028 & 0.627  $\pm$  0.010 & 0.613  $\pm$  0.025  & 0.811  $\pm$ 0.087 &
 0.481  $\pm$ 0.024 &
 0.748  $\pm$ 0.062 &
 0.911  $\pm$ 0.038 &
 0.521  $\pm$ 0.022 &
 0.800  $\pm$ 0.010
\\

    3dSS \cite{cox1996afni} & FLIRT \cite{jenkinson2001global} & DW~\cite{jaderberg2015spatial} & 0.902  $\pm$  0.032 & 0.627  $\pm$ 0.010 & 0.601  $\pm$ 0.017 & 0.849  $\pm$ 0.037 &
 0.500  $\pm$ 0.014 &
 0.791  $\pm$ 0.034 &
   0.869  $\pm$ 0.039 &
0.508  $\pm$ 0.023  &
0.788  $\pm$ 0.021
\\

    BSE \cite{shattuck2002brainsuite} & FLIRT \cite{jenkinson2001global} &DW~\cite{jaderberg2015spatial} &0.938  $\pm$ 0.022
& 0.668  $\pm$ 0.010 &
0.620  $\pm$ 0.008  & 0.846  $\pm$ 0.112 &
0.518  $\pm$ 0.035 &
0.804  $\pm$ 0.019 &
 0.873  $\pm$ 0.064 &
0.521  $\pm$ 0.026 &
0.799  $\pm$ 0.015
 \\

    \midrule

    BET \cite{smith2002fast} & ANTs \cite{avants2009advanced} & DW~\cite{jaderberg2015spatial} & 0.935  $\pm$ 0.028 & 0.630  $\pm$ 0.013 &
0.625  $\pm$ 0.010  & 0.811  $\pm$ 0.087 &
0.476  $\pm$ 0.027 &
0.744  $\pm$ 0.062 &
 0.911  $\pm$ 0.038 &
0.524  $\pm$ 0.022 &
0.792  $\pm$ 0.018

 \\

    3dSS \cite{cox1996afni} & ANTs \cite{avants2009advanced} & DW~\cite{jaderberg2015spatial} & 0.902  $\pm$ 0.032 &
0.632  $\pm$ 0.011 &
0.524  $\pm$ 0.133  & 0.849  $\pm$ 0.037 &
0.498  $\pm$ 0.016 &
0.758  $\pm$ 0.041 &
 0.869  $\pm$ 0.039 &
0.508  $\pm$ 0.026 &
0.762  $\pm$ 0.030\\

    BSE \cite{shattuck2002brainsuite} & ANTs \cite{avants2009advanced} &DW~\cite{jaderberg2015spatial} & 0.938  $\pm$ 0.022 &
0.671  $\pm$ 0.008 &
0.614  $\pm$ 0.023  & 0.846  $\pm$ 0.112 &
0.524  $\pm$ 0.035 &
0.807  $\pm$ 0.012 &
 0.873  $\pm$ 0.064 &
0.522  $\pm$ 0.027 &
0.787  $\pm$ 0.016\\
    \midrule

        BET \cite{smith2002fast} & VM \cite{balakrishnan2018unsupervised} & DW~\cite{jaderberg2015spatial} & 0.935  $\pm$ 0.028 &
0.627 $\pm$ 0.016 &
0.607 $\pm$ 0.019  & 0.811  $\pm$ 0.087 &
0.465 $\pm$ 0.034 &
0.804 $\pm$ 0.015 &
 0.911  $\pm$ 0.038 &
0.525 $\pm$ 0.023 &
0.795 $\pm$ 0.013 \\

    3dSS \cite{cox1996afni} & VM \cite{balakrishnan2018unsupervised} & DW~\cite{jaderberg2015spatial} &0.902  $\pm$ 0.032 &
0.623 $\pm$ 0.009 &
0.590 $\pm$ 0.009  & 0.849  $\pm$ 0.037 &
0.496 $\pm$ 0.018 &
0.814 $\pm$ 0.007 &
 0.869  $\pm$ 0.039 &
0.509 $\pm$ 0.027 &
0.789 $\pm$ 0.017\\

    BSE \cite{shattuck2002brainsuite} & VM \cite{balakrishnan2018unsupervised} &DW~\cite{jaderberg2015spatial} & 0.938  $\pm$ 0.022 &
0.670 $\pm$ 0.003 &
0.616 $\pm$ 0.010  & 0.846  $\pm$ 0.112 &
0.515 $\pm$ 0.041 &
0.805 $\pm$ 0.016 &
 0.873  $\pm$ 0.064 &
0.524 $\pm$ 0.028 &
0.795 $\pm$ 0.013\\
    \midrule

    BET \cite{smith2002fast} & CRN \cite{zhao2019recursive} & DW~\cite{jaderberg2015spatial} & 0.935  $\pm$ 0.028 &
0.633 $\pm$ 0.017 &
0.618 $\pm$ 0.022  & 0.811  $\pm$ 0.087 &
0.467 $\pm$ 0.034 &
0.806 $\pm$ 0.015 &
 0.911  $\pm$ 0.038 &
0.527 $\pm$ 0.023 &
0.800 $\pm$ 0.011\\

    3dSS \cite{cox1996afni} & CRN \cite{zhao2019recursive} & DW~\cite{jaderberg2015spatial} & 0.902  $\pm$ 0.032 &
0.630 $\pm$ 0.012 &
0.610 $\pm$ 0.013  & 0.849  $\pm$ 0.037 &
0.498 $\pm$ 0.017 &
0.817 $\pm$ 0.007 &
 0.869  $\pm$ 0.039 &
0.513 $\pm$ 0.028 &
0.794 $\pm$ 0.014\\

    BSE \cite{shattuck2002brainsuite} & CRN \cite{zhao2019recursive} &DW~\cite{jaderberg2015spatial} & 0.938  $\pm$ 0.022 &
0.674 $\pm$ 0.006 &
0.626 $\pm$ 0.010  & 0.846  $\pm$ 0.112 &
0.518 $\pm$ 0.040 &
0.809 $\pm$ 0.017 &
 0.873  $\pm$ 0.064 &
0.527 $\pm$ 0.028 &
0.796 $\pm$ 0.013\\
    \midrule

        BET \cite{smith2002fast} & \multicolumn{2}{c}{DeepAtlas \cite{xu2019deepatlas}} & 0.935  $\pm$ 0.028 & 0.627 $\pm$ 0.016 & 0.645 $\pm$ 0.009  & 0.811  $\pm$ 0.087 &
0.467 $\pm$ 0.033 &
0.814 $\pm$ 0.017 &
 0.911  $\pm$ 0.038 &
0.524 $\pm$ 0.024 &
0.811 $\pm$ 0.011 \\

    3dSS \cite{cox1996afni} & \multicolumn{2}{c}{DeepAtlas \cite{xu2019deepatlas}} & 0.902  $\pm$ 0.032 & 0.625 $\pm$ 0.013 & 0.640 $\pm$ 0.010  & 0.849  $\pm$ 0.037 &
0.498 $\pm$ 0.018 &
0.828 $\pm$ 0.007 &
 0.869  $\pm$ 0.039 &
0.509 $\pm$ 0.027 &
0.808 $\pm$ 0.013\\

    BSE \cite{shattuck2002brainsuite} & \multicolumn{2}{c}{DeepAtlas \cite{xu2019deepatlas}} & 0.938  $\pm$ 0.022 &
0.667 $\pm$ 0.006 &
0.648 $\pm$ 0.009  & 0.846  $\pm$ 0.112 &
0.518 $\pm$ 0.041 &
0.817 $\pm$ 0.017 &
 0.873  $\pm$ 0.064 &
0.525 $\pm$ 0.029 &
0.807 $\pm$ 0.014\\
    \midrule

    \multicolumn{2}{c}{ERNet \cite{su2022ernet}} & DW~\cite{jaderberg2015spatial} & 0.942 $\pm$ 0.010 &
0.675 $\pm$ 0.004 &
0.620 $\pm$ 0.012  & 0.935 $\pm$ 0.006 &
0.573 $\pm$ 0.007 &
0.818 $\pm$ 0.007 &
 0.915 $\pm$ 0.019 &
0.544 $\pm$ 0.029 &
0.799 $\pm$ 0.012\\
    \midrule
    \multicolumn{3}{c}{\textbf{JERS (ours)}} &
   \textbf{ 0.944 $\pm$ 0.008} &
\textbf{0.679 $\pm$ 0.005} &
\textbf{0.651 $\pm$ 0.011}  & \textbf{0.937 $\pm$ 0.005} &
\textbf{0.574 $\pm$ 0.006} &
\textbf{0.840 $\pm$ 0.010} &
 \textbf{0.917 $\pm$ 0.019} &
\textbf{0.550 $\pm$ 0.025} &
\textbf{0.832 $\pm$ 0.010}
\\
    
    \bottomrule    
    \end{tabular}}
    \vspace{-10pt}
\end{table*}

\subsection{Experimental Results}
Our JERS is compared with baseline methods on extraction, registration and registration accuracy. We measure the performance of each task with its corresponding metrics, and also record the time taken for each baseline method. Based on the experimental results, we find that JERS not only consistently outperforms other alternatives in terms of extraction, registration, and segmentation, but also exhibits superior time efficiency and robustness.

\subsubsection{Evaluation Metrics.}

To evaluate the extraction and segmentation accuracy, we measure the volume overlap between the predicted and ground-truth masks. To evaluate the registration performance, we calculate the mutual information between the warped image and the target image. Details can be found in Appendix~\ref{section: appendix evaluation Metrics}.

\subsubsection{Experiment Setting.}
We divide the datasets into training, validation and test sets. The training set is utilized for parameter learning of the model, while the validation set is employed to evaluate the performance of hyperparameter settings (\eg the weight of the segmentation loss term). The test set is used only once to report the final evaluation results for each model. It should be noted that the IBSR dataset is exclusively used for testing purposes. We describe the detail of the data processing, JERS settings and baseline settings in Appendix~\ref{section: appendix Data Preprocessing}, \ref{section: appendix jers settings} and \ref{section: appendix baselines settings}.
The source code is available at~\url{https://github.com/Anonymous4545/JERS}.

\definecolor{mypgreen}{RGB}{142, 179, 142}
\definecolor{mypred}{RGB}{181, 83, 53}
\definecolor{myppurple}{RGB}{175, 145, 185}
\definecolor{mypyellow}{RGB}{206,226,46}

\begin{figure*}[t]
  \centering
  \includegraphics[width=\linewidth]{fig/main_result.pdf}
  \vspace{-15pt}
  \caption{Visual comparisons for brain extraction, registration and segmentation tasks. We render a 3D visualization of the image and display the middle slice in three different planes: sagittal, axial and coronal. The left side contains the source and target (template) images and their corresponding ground truth labels. We show the extraction, registration and segmentation results of each method and its corresponding predictive labels used for performance evaluations. For the extraction task, a predicted extraction mask (marked by green color) should coincide as much as possible with the ground truth extraction mask of the source image. Likewise, in the segmentation task, a predicted segmentation mask (marked by different color regions) should well-overlap with the ground-truth segmentation mask of the source image. For the registration task, the higher the similarity of the registered brain to the template brain, the better.
  }
  \label{fig:main result}
  \vspace{-15pt}
\end{figure*}

\subsubsection{Extraction, Registration and Segmentation Results.} Table~\ref{tab:main res} presents the results for the compared methods as well as the proposed JERS in extraction, registration, and segmentation tasks. Through a comprehensive evaluation across three datasets, JERS outperforms existing methods in all metrics.

For the extraction task, we observed that the joint-based extraction methods (JERS and ERNet) outperform other single-stage extraction methods, especially on the
CC359 dataset. Specifically, we observed a gain in extraction dice score up to $10.4\%$ compared to the best single-stage extraction method 3dSkullStrip. 
Furthermore, joint-based extraction methods prove to be more robust compared to other alternatives, given their steady performance and achievement of the lowest standard deviation across all datasets.
%Besides, joint-based extraction methods are more robust in the extraction task than other alternatives, as it performs consistently well and obtains the smallest standard deviation across all datasets.

When observing registration performance, joint-based registration methods (JERS and ERNet) also outperform all other methods across all datasets.
Significantly, our findings suggest that the registration performance of most methods is constrained by the result of its corresponding extraction method. This underscores the fact that the precision of extraction considerably influences the subsequent registration task's quality. Joint-based registration methods leverage this characteristic to yield enhanced results through collaborative learning.
%Notably, we find that the registration result of almost every method is bounded by the result of its corresponding extraction method.
%This proves that the accuracy of extraction significantly impacts the quality of the subsequent registration task.
%Joint-based registration methods capture this property to deliver an improved result via collective learning.

For the segmentation task, once again, we find that joint-based segmentation methods (JERS and DeepAtlas) are superior to DW. This demonstrates that joint learning with the registration task can help the segmentation task to boost its performance.

Overall, joint-based methods (JERS, ERNet and DeepAtlas) outperform other pipeline-based methods in their respective joint tasks. However, the partially joint methods perform poorly on their stand-alone task. ERNet performs well on extraction and registration but inferior on the segmentation task. Similarly, although DeepAtlas achieves good results on the segmentation task, the stand-alone extraction method limits its extraction and registration performance. Benefiting from fully end-to-end joint learning, only JERS can perform well in all tasks.

\subsubsection{Qualitative Analysis.}
In Figure~\ref{fig:main result}, 
we visually compare the performance of our JERS and other approaches on the LPBA test set.
Upon observation, it is evident that JERS achieves more accurate brain extraction compared to BET, 3dSkullStrip, and BSE. The brain extraction mask predicted by JERS overlaps closely with the ground truth extraction mask, while the masks predicted by other extraction methods include noticeable non-brain tissues.  Regarding registration results, JERS also outperforms the other methods. The final registered image of JERS exhibits a higher resemblance to the target image compared to the alternatives. Importantly, the inaccurate extraction results with non-brain tissue further impact the subsequent registration results and ultimately affect the overall performance. This supports our assertion that failed extraction can propagate errors to the subsequent registration task, resulting in irreparable consequences. In terms of the segmentation task, JERS produces results that closely overlap to the ground truth segmentation mask. Overall, we can see that better extraction leads to better registration, and better registration yields better segmentation.

\subsubsection{Ablation Study.} To demonstrate the effectiveness of our JERS, we compared five variants of JERS in Table~\ref{tab: ablation}. We first freeze the extraction module, the registration module, and the segmentation module of JERS respectively. JERS w/o Ext consistently produces the extraction mask with all values of 1 (\ie no pattern be removed from the source image). JERS w/o Reg only outputs the identity affine matrix representing no displacement applied to the image. JERS w/o Seg directly warps the segmentation mask to the source image space (\ie no segmentation network exists). The results show that the extraction module, the registration module, and the segmentation module are essential to JERS, and removing any of them degrades the performance of all tasks. We then evaluate the effects of multi-stage and extraction mask smoothing design. The results show that they significantly boost the performance of all tasks. Since all tasks are learned in a collective manner, the performance boost of one module is shared by all other modules.

\subsubsection{Running Efficiency.}
We measure the efficiency of JERS by comparing its running time with other baselines. The measurement is made on the same device with an Intel$^{\circledR}$ Xeon$^{\circledR}$ E5-2667 v4 CPU and an NVIDIA Tesla A100 GPU.
As indicated in Table~\ref{tab: time_res}, all joint-based methods are faster than existing three-stage pipeline-based methods. This is because they can efficiently perform their corresponding extraction, registration and segmentation tasks end-to-end on the same device. 
GPU implementations for BET, 3dSkullStrip, BSE, FLIRT, and ANTs are not available~\cite{smith2002fast,cox1996afni,shattuck2002brainsuite,jenkinson2001global,avants2009advanced}.

\subsubsection{Influence of Parameters.}
\begin{figure}[t]
\begin{tikzpicture}
\begin{axis}[
    title = {(a) Extraction Accuracy},
    outer sep=0,
    width=0.57\columnwidth,
    height=0.42\columnwidth,
    xlabel={Number of stages},
x label style={at={(axis description cs:0.5,-0.13)},anchor=north},
y label style={at={(axis description cs:-0.13,.5)},anchor=south},
    % xlabel near ticks,
    % ylabel near ticks,
    ylabel={Dice (Extraction)},
    xmin=-0.3, xmax=6.3,
    ymin=0.9, ymax=0.950,
    xtick={0,1, 2, 3, 4, 5, 6},
    ytick={0.900,0.910,0.920,0.930,0.940},
    yticklabels={0.900,0.910,0.920,0.930,0.940},
    xticklabels={(1,1), (2,2), (3,3), (4,4), (5,5), (6,6), (7,7)},
    ymajorgrids=true,
    grid style=dashed,
    tick label style={font=\tiny},
    label style={font=\tiny},
    title style={font=\footnotesize,align=center, yshift=-0.37\columnwidth},
]

\addplot[
    dashed,
    color=red,
    mark=diamond,
    mark options=solid,
    ]
    coordinates {
    (0,0.9020)(1,0.9143)(2,0.9256)(3,0.9307)(4,0.944)(5,0.943)(6,0.9441)
    };

\end{axis}
\end{tikzpicture}
\begin{tikzpicture}
\begin{axis}[
    title = {(b) Registration Accuracy},
    outer sep=0,
    width=0.56\columnwidth,
    height=0.42\columnwidth,
    xlabel={Number of stages},
x label style={at={(axis description cs:0.5,-0.13)},anchor=north},
y label style={at={(axis description cs:-0.13,.5)},anchor=south},
    % xlabel near ticks,
    % ylabel near ticks,
    ylabel={MI (Registration)},
    xmin=-0.3, xmax=6.30,
    ymin=0.660, ymax=0.685,
    xtick={0,1,2,3,4,5,6},
    ytick={0.660,0.665, 0.670, 0.675, 0.680, 0.685},
    yticklabels={0.660, 0.665, 0.670, 0.675, 0.680, 0.685},
    xticklabels={(1,1), (2,2), (3,3), (4,4), (5,5), (6,6), (7,7)},
    ymajorgrids=true,
    grid style=dashed,
    tick label style={font=\tiny},
    label style={font=\tiny},
    title style={font=\footnotesize,align=center, yshift=-0.37\columnwidth},
]

\addplot[
    dashed,
    color=blue,
    mark=triangle,
    mark options=solid,
    ]
    coordinates {
    (0,0.6615)(1,0.6643)(2,0.6666)(3,0.6753)(4,0.679)(5,0.6804)(6,0.6806)
    };

\end{axis}
\end{tikzpicture}
\begin{tikzpicture}
\begin{axis}[
    title = {(c) Segmentation Accuracy},
    outer sep=0,
    width=0.57\columnwidth,
    height=0.42\columnwidth,
    xlabel={Number of stages},
x label style={at={(axis description cs:0.5,-0.13)},anchor=north},
y label style={at={(axis description cs:-0.13,.5)},anchor=south},
    % xlabel near ticks,
    % ylabel near ticks,
    ylabel={Dice (Segmentation)},
    xmin=-0.3, xmax=6.30,
    ymin=0.600, ymax=0.655,
    xtick={0,1,2,3,4,5,6},
    ytick={0.605,0.615, 0.625, 0.635, 0.645, 0.655},
    yticklabels={0.605,0.615, 0.625, 0.635, 0.645, 0.655},
    xticklabels={(1,1), (2,2), (3,3), (4,4), (5,5), (6,6), (7,7)},
    ymajorgrids=true,
    grid style=dashed,
    tick label style={font=\tiny},
    label style={font=\tiny},
    title style={font=\footnotesize,align=center, yshift=-0.37\columnwidth},
]

\addplot[
    dashed,
    color=green!50!black,
    mark=o,
    mark options=solid,
    ]
    coordinates {
    (0,0.6089)(1,0.6118)(2,0.6239)(3,0.6408)(4,0.651)(5,0.6490)(6,0.6494)
    };

\end{axis}
\end{tikzpicture}
\begin{tikzpicture}
\begin{semilogxaxis}[
    %log ticks with fixed point,
    % extra x tick style={
    % log identify minor tick positions=true,
    % },
    % inner sep=0,
    title = {(d) Varying of $\lambda$},
    outer sep=0,
    width=0.56\columnwidth,
    height=0.42\columnwidth,
    xlabel={Segmentation loss weight $\lambda$},
x label style={at={(axis description cs:0.5,-0.13)},anchor=north},
y label style={at={(axis description cs:-0.13,.5)},anchor=south},
    ylabel={Dice (Segmentation)},
    xmin=6e-8, xmax=0.16,
    ymin=0.440, ymax=0.680,
    xtick={  1e-7, 1e-6, 1e-5, 1e-4,1e-3, 1e-2, 1e-1  },
    xticklabels={  $10^{-7}$, $10^{-6}$, $10^{-5}$, $10^{-4}$,$10^{-3}$, $10^{-2}$, },
    ytick={0.440, 0.480, 0.520, 0.560, 0.600,0.640, 0.680},
    yticklabels={0.440, 0.480, 0.520, 0.560, 0.600,0.640, 0.680},
    ymajorgrids=true,
    grid style=dashed,
    tick label style={font=\tiny},
    label style={font=\tiny},
    title style={font=\footnotesize,align=right, yshift=-0.37\columnwidth},
]

\addplot[
    dashed,
    color=orange,
    mark=square,
    mark options=solid,
    ]
    coordinates {(1e-7, 0.4465)
    (1e-6, 0.5410)(1e-5,0.6266)(1e-4,0.6335)(1e-3,0.6369)(1e-2,0.6423)(1e-1,0.6513)
    };

\end{semilogxaxis}
\end{tikzpicture}
    \vspace{-11pt}
    \caption{Effect of varying the number of stages of the JERS and the segmentation loss weight $\lambda$.}
    \label{fig:stage dice}
    \vspace{-11pt}
\end{figure}
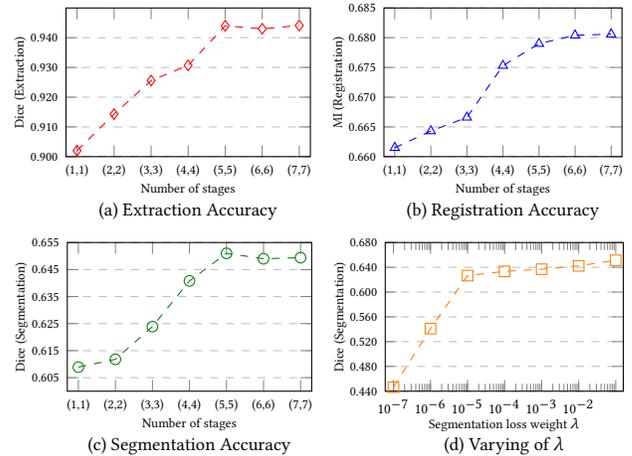

We study two crucial hyperparameters of our JERS: the number of stages for extraction and registration modules and the value of segmentation loss weight~$\lambda$. 
In our multi-stage design, the number of stages in the network represents the depth of the model and the number of iterations for the extraction and registration tasks. Essentially, increasing the number of stages allows for more refinements in the extraction and registration processes.
As illustrated in Figure~\ref{fig:stage dice}(a, b, c), we adjust the number of stages of extraction and registration to study their influence.
The outcomes show that increasing the stages boosts the performance of all tasks, validating the notion that a multi-stage framework results in superior overall performance in a joint learning system.
%The results indicate that the performance of all tasks improves with additional stages.
%This supports the idea that a multi-stage design yields improved overall performance in a joint learning system.

As mentioned in Section~\ref{section:end-to-end training}, we introduce a segmentation loss term to learn a better segmentation network.
To show the effectiveness of the loss term, we vary different values of the loss weight $\lambda$ as shown in Figure~\ref{fig:stage dice}(d).
As the weight of the loss term gradually increases, the segmentation dice score grows as well.
This indicates that our JERS benefits from the segmentation loss term.

% %-----------------------------------------------
% % Related Work Section
\section{Related Work}
\label{sec:related}
\noindent \textbf{Neuroimage extraction.} 
In the past decade, numerous methods have emerged, highlighting the significance of the brain extraction problem. 
Smith et al.~\cite{smith2002fast} introduced a deformable model that fits the brain surface using a locally adaptive set model. 3dSkullStrip~\cite{cox1996afni} is a modified version of~\cite{smith2002fast} that employs points outside the brain surface to guide mesh evolution. Shattuck et al.~\cite{shattuck2002brainsuite} utilized anisotropic diffusion filtering and a 2D Marr Hildreth edge detector for brain boundary identification.
However, these methods heavily rely on parameter tuning and manual quality control, which are time-consuming and labor-intensive. Recently, brain extraction has benefited from the introduction of deep learning approaches, which exhibit exceptional performance and speed. 
Kleesiek et al.~\cite{kleesiek2016deep} proposed a voxel-wise 3D CNN for skull stripping, while Hwang et al.~\cite{hwang20193d} demonstrated the effectiveness of 3D-UNet in achieving competitive results. However, these learning-based approaches often necessitate a substantial amount of properly labeled data for effective training, which is a challenge considering that neuroimage datasets are typically small and costly to annotate.

\noindent \textbf{Neuroimage registration.}
Conventional techniques for image registration~\cite{avants2009advanced, avants2008symmetric, jenkinson2001global} typically aim to maximize image similarity by iteratively optimizing transformation parameters. Commonly used intensity-based similarity measures include normalized cross-correlation (NCC) and mutual information (MI), among others. However, this iterative optimization approach often suffers from high computational costs and being stuck in local optima, resulting in inefficient and unreliable registration outcomes. Recently, numerous deep learning-based methods have been proposed, offering improved computational efficiency and registration performance. For instance, Sokooti et al.~\cite{sokooti2017nonrigid} introduced a multi-scale 3D CNN called RegNet, which learns the displacement vector field (DVF) for 3D chest CT registration. Although these methods demonstrate competitive results, they require supervision. To overcome this limitation, unsupervised registration methods~\cite{balakrishnan2018unsupervised,zhao2019recursive} have garnered significant attention and shown promising outcomes.

\begin{table}[t]
    \centering
    \caption{Ablation studies of JERS on LPBA40 dataset}
    \label{tab: ablation}
    \vspace{-10pt}
    \resizebox{0.85\linewidth}{!}{
    \begin{tabular}{lccrcc}
    \toprule

    Methods & Dice$_{\mathrm{ext}}$ $\uparrow$ & MI$_{\mathrm{reg}}$ $\uparrow$ & Dice$_{\mathrm{seg}}$ $\uparrow$ \\
    
\midrule

    JERS w/o Ext &  0.216 $\pm$ 0.018 &
0.579 $\pm$ 0.017 &
0.224 $\pm$ 0.103 \\
 JERS w/o Reg  &
0.311 $\pm$ 0.030 &
0.217 $\pm$ 0.036 &
0.268 $\pm$ 0.157 \\

 JERS w/o Seg   &
0.942 $\pm$ 0.009 &
0.677 $\pm$ 0.005 &
0.620 $\pm$ 0.008 \\

 JERS w/o Multi-stage &
0.902 $\pm$ 0.005 &
0.662 $\pm$ 0.003 &
0.609 $\pm$ 0.017 \\

 JERS w/o Ext smoothing  &
0.931 $\pm$ 0.006  &
0.673 $\pm$ 0.003  &
0.649 $\pm$ 0.008 \\
\midrule
 \textbf{JERS}   &
\textbf{0.944  $\pm$  0.008} &
\textbf{0.679  $\pm$  0.005} &
\textbf{0.651  $\pm$  0.011} \\

    \bottomrule    
    \end{tabular}}
    \vspace{-10pt}
\end{table}

\begin{table}[t]
    \centering
    \caption{Running Time of compared methods on LPBA40 dataset.}
    \label{tab: time_res}
    \vspace{-10pt}
    \resizebox{0.65\linewidth}{!}{
    \begin{tabular}{lccrcc}
    \toprule
        \multicolumn{3}{c}{Methods}              & \multicolumn{3}{c}{Time (Sec) $\downarrow$} \\
        \cmidrule(lr){1-3}\cmidrule(lr){4-6}

   Ext & Reg  & Seg & Ext & Reg & Seg \\
\midrule
       BET \cite{smith2002fast} & FLIRT \cite{jenkinson2001global} & DW~\cite{jaderberg2015spatial} &  2.4751 &
4.6857 &
0.1457 \\

    3dSS \cite{cox1996afni} & ANTs \cite{avants2009advanced} & DW~\cite{jaderberg2015spatial} &  176.9355 &
2.6705&
0.0370 \\

\midrule

    BSE \cite{shattuck2002brainsuite} & VM \cite{balakrishnan2018unsupervised} &DW~\cite{jaderberg2015spatial} &  3.5119 &
0.0050 &
0.0001 \\

    BET \cite{smith2002fast} & CRN \cite{zhao2019recursive} & DW~\cite{jaderberg2015spatial} &  2.4751 &
0.0151 &
0.0001 \\

\midrule

        BET \cite{smith2002fast} & \multicolumn{2}{c}{DeepAtlas \cite{xu2019deepatlas}} &  2.4751 &
\multicolumn{2}{c}{0.0018}  \\

\midrule

    \multicolumn{2}{c}{ERNet \cite{su2022ernet}} & DW~\cite{jaderberg2015spatial} &  \multicolumn{2}{c}{0.0420} &
0.0001 \\

\midrule

\multicolumn{3}{c}{JERS (ours)} & \multicolumn{3}{c}{0.0455
} \\

    \bottomrule    
    \end{tabular}}
    \vspace{-12pt}
\end{table}

\noindent \textbf{Neuroimage segmentation.}
CNN-based approaches have demonstrated superior performance in terms of speed and accuracy for supervised neuroimage segmentation. Based on the dimensionality of network operation, current work mainly falls into two categories:  2D CNN-based network and 3D CNN-based network. 2D CNN-based network processes the volumetric neuroimage data slice by slice, and assembles the final segmentation mask by putting segmentation results on all the slices together. The most representative networks include UNet \cite{ronneberger2015u} and Attention-UNet \cite{oktay2018attention}. Since the 2D CNN-based network treats each slice separately, the spatial information encoded between slices is not utilized, negatively affecting the segmentation performance. This limitation prompts the use of the 3D CNN-based network for 3D segmentation. The most representative network among all is the 3D UNet \cite{cciccek20163d}.
%a 3D version of the original UNet architecture.
Despite the success CNN-based approaches had in neuroimage segmentation, those approaches still possess a data-hungry nature and require a large number of labeled neuroimages for training, which is expensive to acquire.  

\noindent \textbf{Joint neuroimage registration and segmentation}
The neuroimage registration and segmentation tasks are deeply co-related and mutually facilitating, thus should not be treated separately. Specifically, the anatomy structure produced by neuroimage segmentation can provide auxiliary information for neuroimage segmentation. This idea is practiced in \cite{hu2018weakly} and demonstrated to be effective. Complementary, the registration task can also aid the segmentation task, typically studied in the scope of atlas-based segmentation and data augmentation. In atlas-based segmentation, the atlas label is transferred to the unlabeled image space using the geometric transformation estimated by a registration module \cite{wang2020lt, dinsdale2019spatial}. Taking the idea a step further, for data augmentation, new image-label pairs are generated by means of sampling the geometric and style transformation \cite{ding2021modeling, zhao2019data}. The above methods still focus on a single task. To address this limitation, exploring work has been conducted to combine registration and segmentation together, by optimizing them either jointly \cite{qiu2021u} or alternatively \cite{xu2019deepatlas,he2020deep}, and is able to obtain better results.

% %-----------------------------------------------
% % Conclusion Section
\section{Conclusion}
\label{sec:con} 

This paper introduces a novel unified framework, called JERS, for one-shot joint extraction, registration and segmentation.
In contrast to prior research, our proposed method seamlessly integrates the three tasks into a single system, enabling joint optimization.
Specifically, JERS contains a group of collective extraction, registration and segmentation modules.
These three modules help each other boost extraction, registration, and segmentation performance with only one labeled template available.
Furthermore, our method facilitates incremental progress for each task, thereby enhancing the overall performance to a greater extent.
The experimental results substantiate that JERS not only surpasses state-of-the-art approaches in terms of extraction, registration, and segmentation accuracy but also exhibits high robustness and time-efficiency.

\section{Acknowledgments}
\label{sec:ack}
This work is partially supported by the National Science Foundation (MRI 2215789 and IIS 1909879) and Lehigh's grants (S00010293 and 001250).

%\newpage
% balance the two columns of the last page
\balance

\bibliographystyle{ACM-Reference-Format}
\bibliography{08_reference_short}

\newpage

\appendix

\section{Appendix}
\label{sec:appendix}

This section provides more details of evaluation metrics and experiment settings to support the reproducibility of the results in this paper. Our code and data have been made publicly available at~\url{https://github.com/Anonymous4545/JERS}.

\subsection{Evaluation Metrics}
\label{section: appendix evaluation Metrics}
Our defined problem aims to solve the brain extraction, registration and segmentation tasks simultaneously: 1) identify the brain region (\ie whole cerebral tissue) within the source image; 2) align the extracted cerebral tissues to the target image; 3) identify the anatomical segmentation within the source image. Thus, we evaluate the accuracy of extraction, registration and segmentation to show the performance of our proposed method and compared methods.

\subsubsection{Extraction Performance.}
The brain MRI datasets incorporate the ground truth of the brain mask, representing the labeling of brain tissue in the source image. To assess the accuracy of extraction, we evaluate the volume overlap of brain masks using the Dice score, which can be expressed as:
%The brain MRI datasets contain the brain mask ground truth, which is the label of brain tissue in the source image. To evaluate the extraction accuracy, we measure the volume overlap of brain masks by Dice score, which can be formulated as:
\begin{equation}
\text{Dice}_{\mathrm{ext}}=2 \cdot \frac{|\mathbf{\hat{M}} \cap \mathbf{M}|}{|\mathbf{\hat{M}}|+|\mathbf{M}|},
\end{equation}
where $\mathbf{\hat{M}}$ denotes the predicted brain mask and $\mathbf{M}$ is the corresponding ground truth. If $\mathbf{\hat{M}}$ signifies a precise extraction, we anticipate a high degree of overlap between the non-zero areas in $\mathbf{\hat{M}}$ and $\mathbf{M}$.

\subsubsection{Registration Performance.} 
To evaluate the registration performance, we calculate the mutual information \cite{maes2003medical, thevenaz2000optimization,pluim2000image,563664} between the warped image (\ie registered) $\mathbf{W}$ and the target image~$\mathbf{T}$:
\begin{equation}
    MI(\mathbf{W}, \mathbf{T}) = \sum_{w, t} p_{\mathbf{W} \mathbf{T}}(w, t)\log \frac{p_{\mathbf{W} \mathbf{T}}(w, t)}{p_{\mathbf{W}}(w) \cdot p_{\mathbf{T}}(t)}
\end{equation}
where $p_{\mathbf{W}}(w)$ and $p_{\mathbf{T}}(t)$ are the marginal probability distributions of image $\mathbf{W}$ and $\mathbf{T}$, respectively. $p_{\mathbf{W} \mathbf{T}}(w, t)$ is the joint probability distribution. The mutual information measures the mutual dependence between $\mathbf{W}$ and $\mathbf{T}$. If the warped image $\mathbf{W}$ and the target image $\mathbf{T}$ are geometrically aligned, we expect the mutual information to be maximal.

\subsubsection{Segmentation Performance.}
We evaluate the segmentation accuracy by measuring the volume overlap of anatomical segmentation, which are the location labels of different tissues in the brain MRI image. If the segmentation task performs well,  the predicted segmentation mask should overlap with the ground truth. Similar to the extraction evaluation, we use Dice score to evaluate the overlap of the segmentation masks. A Dice score of 1 signifies that the corresponding structures overlap with the ground truth, whereas a score of 0 denotes the complete absence of overlap. If the image includes multiple labeled anatomical structures, the final score is calculated as the average of the Dice scores for each structure.

\subsection{Details of Data Preprocessing}
\label{section: appendix Data Preprocessing}
The proposed method and baselines are evaluated on three different public brain MRI datasets, LPBA40, CC-359 and IBSR. 

\noindent\textbullet\ \textit{LONI Probabilistic Brain Atlas (LPBA40)}~\cite{shattuck2008construction}: 
The dataset comprises 40 raw T1-weighted 3D brain MRI scans, each from a different patient. It includes corresponding brain masks and segmentation ground truth for 56 anatomical structures. The brain mask is used for evaluating the extraction accuracy, while the anatomical segmentations are used for evaluating the segmentation accuracy. Similar to~\cite{balakrishnan2018unsupervised, zhao2019recursive}, our focus is on atlas-based registration, where the first scan serves as the target image and the remaining scans are aligned to it. Out of the 39 remaining scans, 30 are used for training, 5 for validation, and 4 for testing. All scans are cropped and resized to $96\times96\times96$ dimensions.

\noindent\textbullet\ \textit{Calgary-Campinas-359 (CC-359)}~\cite{souza2018open}: 
The dataset consists of 359 raw T1-weighted 3D brain MRI scans from 359 different patients. Additionally, it includes corresponding brain masks and labeled white matter as ground truth. The brain masks are used to evaluate the accuracy of extraction, while the white matter masks are used for segmentation evaluation. Similar to the LPBA40 dataset, our focus is on atlas-based registration. For CC359, we divide the dataset into training, validation, and test sets, consisting of 298, 30, and 30 scans, respectively. All scans are cropped and resized to $96\times96\times96$.

\noindent\textbullet\ \textit{Internet Brain Segmentation Repository (IBSR)} \cite{rohlfing2011image}: 
The dataset consists of 18 raw T1-weighted 3D brain MRI scans from 18 different patients, accompanied by corresponding segmentation results. The segmentation results are merged to create the brain mask. Given the limited sample size, this dataset is exclusively used for testing the model trained on CC359. Consequently, all 18 scans are aligned with the first scan of CC359. After cropping, all scans are resized to $96\times96\times96$ dimensions.

\subsection{Details Settings of JERS}
\label{section: appendix jers settings}
\textbf{Training settings of JERS.} 
Our experiments are conducted on Red Hat Enterprise Linux 7.3, utilizing an Intel$^{\circledR}$ Xeon$^{\circledR}$ E5-2667 v4 CPU and an NVIDIA Tesla A100 GPU.
The code is implemented in Python 3.7.6, and the neural networks are built using PyTorch 1.7.1.
The implementation also makes use of Numpy 1.21.6, SimpleITK 2.0.2, and Nibabel 3.1.1. 
To overcome GPU memory limitations, we employ batch gradient descent, with each training batch consisting of one image pair. The models are optimized using the Adam optimizer, with a learning rate of $1 \times 10^{-6}$.
We also apply image augmentation techniques, including random translation, rotation, and scaling, to the source images during training. For more details, please refer to Table~\ref{tab:transformation}.

\begin{table}[h]
    \centering
    \vspace{-2pt}
    \caption{Range of random transformation.}
    \label{tab:transformation}
    \vspace{-8pt}
    \resizebox{0.70\linewidth}{!}{
    \begin{tabular}{lccc}
    \toprule
    
    \multicolumn{1}{c}{\multirow{3}{*}{Datasets}} &   \multicolumn{3}{c}{Transformation} \\

    \cmidrule(lr){2-4}
    
        &   Translation  & Rotation       & Scale         \\
        &   (Voxels)  & (Degree)  & (Times)    \\

    \midrule
    LPBA40 & $\pm$ 5 & $\pm$ 5  & 0.98 $\sim$ 1.02\\
    
    CC359 & $\pm$ 3 & $\pm$ 3  & 0.99 $\sim$ 1.01\\

    \bottomrule
    \end{tabular}}
    \vspace{-10pt}
\end{table}

\noindent\textbf{Parameters settings of JERS.} The extraction and registration stages are set to 5 in this work. The segmentation loss parameter $\lambda$ and extraction mask smooth parameter $\eta$ in Eq.~(\ref{eq:loss}) are 0.1 and 1, respectively.
The extraction network contains 10 convolutional layers with 16, 32, 32, 64, 64, 64, 32, 32, 32 and 16 filters. 
The registration network adopt 3D CNNs and fully-connected layers to map the input to the dimension of $1\times12$. It contains 6 convolutional layers with 16, 32, 64, 128, 256 and 512 filters. 
The segmentation network contains 10 convolutional layers with 128, 256, 256, 512, 512, 512, 256, 256, 256 and 128 filters.

\newcommand{\code}[1]{\texttt{#1}}

\subsection{Settings of Baselines}
\label{section: appendix baselines settings}
The settings of baselines are followed by~\cite{su2022ernet} for a fair comparison.
\noindent\textbf{Brain Extraction Tool (BET)} \cite{smith2002fast}: 
This skull stripping method is a component of the FSL (FMRIB Software Library) package. It employs a deformable approach to accurately fit the brain surface by utilizing locally adaptive set models.
The command we use for BET is \code{bet <input> <output> -f 0.5 -g 0 -m}, where \texttt{f} and \texttt{g} are fractional intensity threshold and gradient in fractional intensity threshold, respectively. We set them to default values.

\noindent\textbf{3dSkullStrip}~\cite{cox1996afni}: This modified version of BET (Brain Extraction Tool) is integrated into the AFNI (Analysis of Functional NeuroImages) package. It performs skull stripping by employing the expansion paradigm of the spherical surface. The command we use for 3dSkullStrip is \code{3dSkullStrip -input <input>  -prefix <output> -mask\_vol -fac 1000}. \code{fac} is set to the default value.

\noindent\textbf{Brain Surface Extractor (BSE)}~\cite{shattuck2002brainsuite}: This method extracts the brain region by utilizing morphological operations and edge detection techniques. It incorporates anisotropic diffusion filtering to enhance image quality and a Marr Hildreth edge detector to accurately identify the boundaries of the brain. The command we use for BSE is \code{bse -i <input> -o <output> --mask <mask> -p --trim --auto --timer }. Hyperparameters are set to default values.

\noindent\textbf{FMRIB's Linear Image Registration Tool (FLIRT)}~\cite{jenkinson2001global}: This is a fully automated affine brain image registration tool included in the FSL (FMRIB Software Library) package. It performs the registration process without requiring manual intervention, allowing for the alignment of brain images based on affine transformations. The command we use for FLIRT is \code{flirt -in <source> -ref <target> -out <output> -omat <output parameter> -bins 256 -cost corratio -searchrx -90 90 -searchry -90 90 -searchrz -90 90 -dof 12  -interp trilinear}.

\noindent\textbf{Advanced Normalization Tools (ANTs)} \cite{avants2009advanced}: It is a cutting-edge medical image registration toolkit widely used in the field. In our approach, we employ the affine transformation model and cross-correlation metric provided by ANTs for the registration process. 

\noindent\textbf{VoxelMorph (VM)} \cite{balakrishnan2018unsupervised}: This unsupervised image registration method utilizes a neural network to predict the transformation between images. In order to ensure a fair comparison, we re-implemented the method using an affine transformation. The network architecture consists of 6 convolutional layers with filter sizes of 16, 32, 64, 128, 256, and 512. The deformation regularization ratio is set to 10, ensuring smooth and controlled transformations during the registration process.

\noindent\textbf{Directly Warping (DW)}~\cite{jaderberg2015spatial}:
This operation refers to generating a segmentation mask through the process of registration. Once the registration is completed, the segmentation mask of the target image can be directly warped and transformed into the source image space.

\noindent\textbf{Cascaded Registration Networks (CRN)} \cite{zhao2019recursive}:
This is an unsupervised multi-stage registration method that involves iteratively transforming the source image to align with a target image. 
Same to JERS, the number of stages is set to 5.
Within each stage, we configure the network architecture with 6 convolutional layers using filter sizes of 16, 32, 64, 128, 256, and 512.

\noindent\textbf{DeepAtlas} \cite{xu2019deepatlas}: This is an unsupervised learning method for joint registration and segmentation. For a fair comparison, we configure 6 convolutional layers with 16, 32, 64, 128, 256 and 512 filters for the registration module, and the segmentation network contains 10 convolutional layers with 128, 256, 256, 512, 512, 512, 256, 256, 256 and 128 filters.

\noindent\textbf{ERNet} \cite{su2022ernet}: This is an unsupervised learning method for joint extraction and registration. For a fair comparison, the number of stages is set to 5, and we configure 6 convolutional layers with 16, 32, 64, 128, 256 and 512 filters, and the registration network contains 6 convolutional layers with 16, 32, 64, 128, 256 and 512 filters. 
\end{document}